\documentclass[journal]{IEEEtran}
\usepackage{amsmath,amssymb,amsfonts}
\usepackage{cite}
\usepackage{amsthm}
\usepackage{mathrsfs}
\usepackage{float}
\usepackage{graphicx}
\usepackage{subfigure}
\usepackage{textcomp}
\usepackage{xcolor}
\usepackage{colortbl,booktabs}
\usepackage[ruled,vlined]{algorithm2e} 
\usepackage{times}  %Required
\usepackage{helvet}  %Required
\usepackage{courier}  %Required
\usepackage{soul}
\usepackage{url}  %Required
\usepackage{bm}
\usepackage{algorithmic}

\usepackage{changepage}

\newtheorem{theorem}{\textbf{Theorem}}
\newtheorem{theorem*}{Theorem}

\newtheorem{Definition}{Definition}

\newtheorem{definition*}{Problem}

\usepackage{multirow}
\usepackage{lineno}

\usepackage{ifpdf}
\ifCLASSINFOpdf
\else
\fi
\usepackage{array}
\begin{document}
\title{Bridging the Theoretical Bound and Deep Algorithms for Open Set Domain Adaptation}
% \title{UREDAnet: A Method For Open Set Domain Adaptation With A Risk Estimator And Double Adversarial Net}

\author{Li Zhong$^\dagger$,~Zhen~Fang$^\dagger$, Feng Liu,~\IEEEmembership{Student Member,~IEEE},\\Bo Yuan,~\IEEEmembership{Member,~IEEE},~Guangquan~Zhang,~and~Jie~Lu$^*$,~\IEEEmembership{Fellow,~IEEE}% <-this % stops a space

\thanks{Li Zhong is with Shenzhen International Graduate School, Tsinghua University, Shenzhen, P.R. China, and Centre for Artificial Intelligence, Faulty of Engineering and Information Technology, University of Technology Sydney, Sydney, NSW, 2007, Australia (e-mail: zhongl18@mails.tsinghua.edu.cn; liallen.zhong@uts.edu.au).

Zhen Fang, Feng Liu, Guangquan Zhang, and Jie Lu are with the Centre for Artificial Intelligence, Faulty of Engineering and Information Technology, University of Technology Sydney, Sydney, NSW, 2007, Australia (e-mail: zhen.fang@student.uts.edu.au; feng.liu@uts.edu.au; guangquan.zhang@uts.edu.au; jie.lu@uts.edu.au).

Bo Yuan is with Shenzhen International Graduate School, Tsinghua University, Shenzhen, P.R. China (e-mail: yuanb@sz.tsinghua.edu.cn).

$^\dagger$Equal contribution. $^*$Corresponding author.

% Guangquan Zhang and Jie Lu are with the Centre for Artificial Intelligence, Faulty of Engineering and Information Technology, University of Technology Sydney, Sydney, NSW, 2007, Australia (e-mail: Guangquan.Zhang@uts.edu.au; Jie.Lu@uts.edu.au)
}}

% \thanks{Li Zhong, Bo Yuan are with Shenzhen International Graduate School, Tsinghua University, Shenzhen, P.R. China. Zhen Fang, Feng Liu, Guangquan Zhang and Jie Lu are with the Centre for Artificial Intelligence, Faulty of Engineering and Information Technology, University of Technology Sydney, Sydney, NSW, 2007, Australia, e-mail: zhen.fang@student.uts.edu.au, \{feng.liu;Guangquan.Zhang; Jie.Lu\}@uts.edu.au.}
% <-this % stops a space
% \thanks{J. Doe and J. Doe are with Anonymous University.
% <-this % stops a space

% \markboth{Journal of \LaTeX\ Class Files,~Vol.~14, No.~8, August~2015}%
% {Shell \MakeLowercase{\textit{et al.}}: Bare Demo of IEEEtran.cls for IEEE Journals}

\maketitle

\begin{abstract}
In the \emph{unsupervised open} \emph{set} \emph{domain} \emph{adaptation} (UOSDA), the target domain contains unknown classes that are not observed in the source domain. Researchers in this area aim to train a classifier to accurately: 1) recognize \emph{unknown target data} (data with unknown classes) and, 2) classify other target data.
To achieve this aim, previous study has proven an upper bound of the target-domain risk, and the \emph{open set difference}, as an important term in the upper bound, is used to measure the risk on unknown target data.
By minimizing the upper bound, a \emph{shallow} classifier can be trained to achieve the aim. 
However, if the classifier is very flexible (\textit{e.g.}, \emph{deep neural networks} (DNNs)), the open set difference will converge to a negative value when minimizing the upper bound, which causes an issue where most target data are recognized as unknown data.
To address this issue, we propose a new upper bound of target-domain risk for UOSDA, which includes four terms: source-domain risk, \emph{$\epsilon$-open set difference} ($\Delta_\epsilon $), distributional discrepancy between domains, and a constant.
Compared to the open set difference, $\Delta_\epsilon $ is more robust against the issue when it is being minimized, and thus we are able to use very flexible classifiers (\textit{i.e.}, DNNs). 
Then, we propose a new principle-guided \emph{deep} UOSDA method that trains DNNs via minimizing the new upper bound. 
Specifically, source-domain risk and $\Delta_\epsilon $ are minimized by gradient descent, and the distributional discrepancy is minimized via a novel open-set conditional adversarial training strategy.
Finally, compared to existing shallow and deep UOSDA methods, our method shows the state-of-the-art performance on several benchmark datasets, including digit recognition (\emph{MNIST}, \emph{SVHN}, \emph{USPS}), object recognition (\emph{Office-31}, \emph{Office-Home}), and face recognition (\emph{PIE}).

\end{abstract}

\begin{IEEEkeywords}
Transfer Learning, Machine Learning, Domain Adaptation, Open Set Recognition.
\end{IEEEkeywords}

% \fi
\IEEEpeerreviewmaketitle

\section{Introduction}
% Introduce DA and UDA
\emph{Domain Adaptation} (DA) methods aim to train a target-domain classifier with data in source and target domains \cite{lu2015transfer}. Based on the variety of data in the target domain (\textit{i.e.}, fully-labeled, partially-labeled, and unlabeled), DA consists of three categories: supervised DA \cite{motiian2017unified, zuo2018fuzzy01, zuo2017granular}, semi-supervised DA \cite{pereira2018semi, saito2019semi,zuo2018fuzzy02}, and \emph{unsupervised DA} (UDA) \cite{liu2017heterogeneous,fang2019unsupervised}. In practice, UDA methods have been deployed to solve diverse real-world problems, such as object recognition \cite{gopalan2011domain, kan2014domain}, cross-domain recommendation  \cite{zhang2017cross}, and sentiment analysis \cite{liu2020heterogeneous}.
% tan2009adapting

\begin{figure}
    \centering
    \includegraphics[scale=0.55, trim=80 160 145 0, clip]{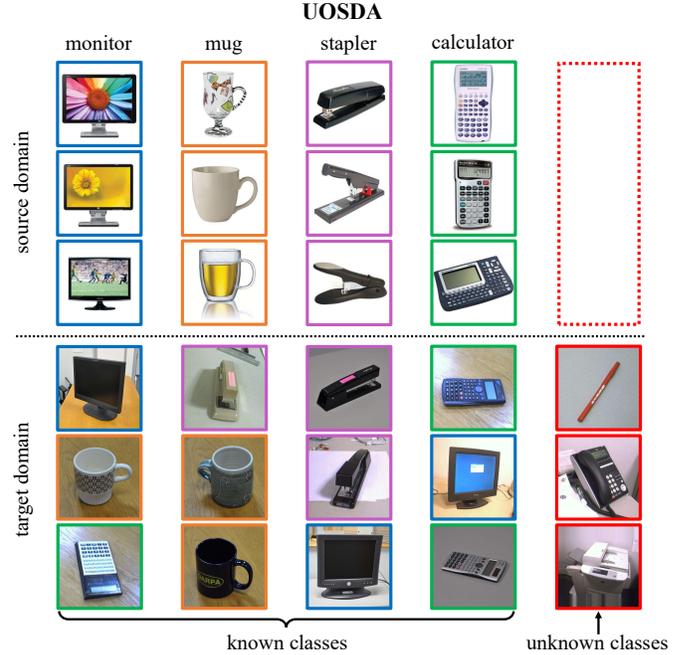}
    \caption{Unsupervised open set domain adaptation (UOSDA). When the target domain does not contain unknown classes, UOSDA will degenerate into the unsupervised closed set domain adaptation (UCSDA).}
    \label{fig:UOSDA}
\end{figure}

% Introduce UCSDA and UOSDA and Why UOSDA is more general in real world (add an example)
There are two common settings in UDA: \emph{unsupervised closed set domain adaptation} (UCSDA) and \emph{unsupervised open set domain adaptation} (UOSDA). UCSDA is a classical scenario in which source and target domains share the same label sets. By contrast, in UOSDA, the target domain contains some unknown classes that are not observed in the source domain, and the data with unknown classes are called \emph{unknown target data}. In Fig. \ref{fig:UOSDA}, the source domain contains four known classes (\textit{i.e.}, monitor, mug, staple, and calculator), but the target domain contains some unknown classes in addition to the classes in the source domain.

UOSDA is more general than UCSDA, since the label sets are usually not consistent between source and target domains in a real-world scenario. Namely, the target domain may contain classes that are not observed in the source domain. For example, a classifier trained with images of various kinds of cats is likely to encounter the image of a dog or another animal in reality. In this case, the UCSDA methods are unable to distinguish the unseen animals (\textit{i.e.,} unknown classes). UOSDA methods, however, can establish a boundary between known classes and unknown classes.

Panareda \textit{et al.}  \cite{panareda2017open} are the first to propose the setting of UOSDA, but the source domain also contains some unknown classes in Panareda's paper. Since it is expensive and prohibitive to obtain data labeled by unknown classes in the source domain, Saito \textit{et al.} \cite{saito2018open} propose a new UOSDA setting where the source domain only contains known classes. In this paper, we focus on the same setting as Saito's paper, which is more realistic \cite{saito2018open,fang2019open}. 

% why UOSDA difficult
In UOSDA, we aim to train a target-domain classifier with labeled data in the source domain and unlabeled data in the target domain. The trained classifier is expected to accurately 1) recognize unknown target data, and 2) classify other target data.
Existing UOSDA methods can be divided into two groups: shallow methods and deep methods. For shallow methods, a recent work \cite{fang2019open} proved an upper bound of target-domain risk, which can provide a theoretical guarantee for the design of a shallow UOSDA method.
For deep methods, since \cite{long2013transfer, yosinski2014transferable, DBLP:conf/icml/DonahueJVHZTD14} have shown that DNNs can learn more transferable features, researchers  presented DNNs-based methods to address the UOSDA problem \cite{saito2018open, feng2019attract, liu2019separate}. Nevertheless, these deep UOSDA methods lack theoretical guarantees. Thus, how to bridge theoretical bound and deep algorithms is both necessary and important for addressing the UOSDA problem.

% Whereas it is intricate to recognize unknown samples from target domain since lacking knowledge about unknown classes. Moreover, distribution alignment is more difficult on account of the existence of unknown classes. If unknown samples are not recognized well before distribution alignment, it is likely to cause negative transfer during the procession of domain adaptation \cite{liu2019separate}.

% zhen's paper
% Zhen \textit{et al.} \cite{fang2019open} introduce an existing theory-based UOSDA method -- a shadow method. In the end, use one sentence to say the drawback of Zhen's method (shadow itself)
In order to train an effective target-domain classifier, Zhen \textit{et al.} \cite{fang2019open} have proven an upper bound of target-domain risk (Eq. (\ref{eq:osd})) for the UOSDA problem and propose a \emph{shallow} UOSDA method. Specifically, the bound consists of four terms: source-domain risk, distributional discrepancy between domains, \emph{open set difference} ($\Delta$), and a constant. Open set difference, as an important term in upper bound, is leveraged to measure the risk of a classifier on unknown target data. 
% There are two terms in open set difference. 
% The first term is the risk of classifying target data as unknown classes and the second term is the risk of classifying source data as unknown classes. The value of open set difference is the first term minus the second term. 
The shallow method in \cite{fang2019open} trains a target-domain classifier by minimizing the empirical estimation of the upper bound.

\begin{figure}[!t]
    \centering
    \includegraphics[scale=0.4,trim=5 10 0 0, clip]{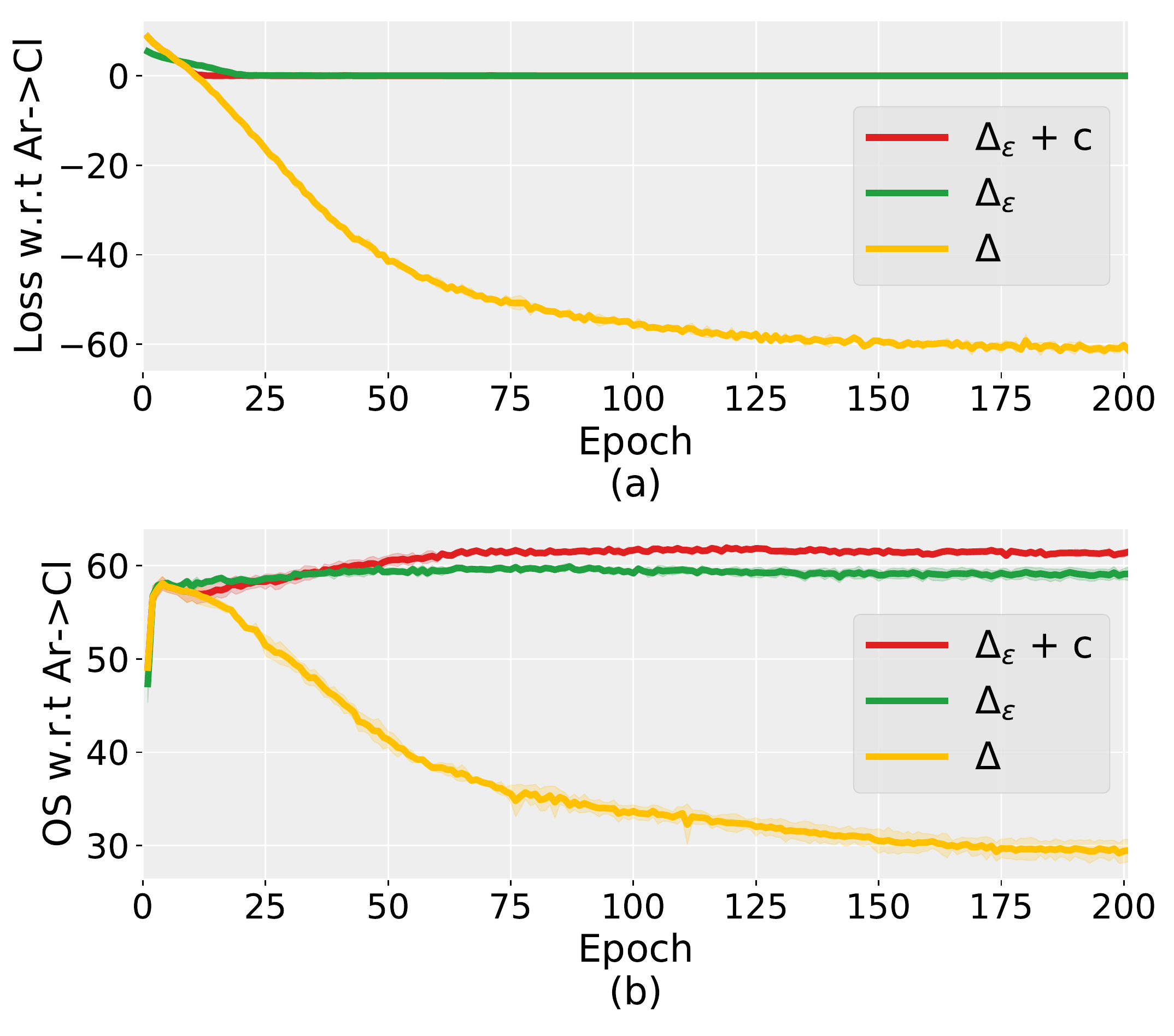}
    \caption{The accuracy of {OS}  and loss \textit{w.r.t.} the task Ar $\rightarrow$ Cl. ``c" denotes the conditional adversarial training strategy. $\Delta_\epsilon $ is the $\epsilon$-open set difference proposed in this paper and $\Delta$ is the open set difference proposed in \cite{fang2019open}. The loss in (a) is the value of $\Delta$ or $\Delta_\epsilon$. It is worth noting that the green line and the red line in (a) are partially  coincident. Here, $\epsilon $ is set as $0$.}
    \label{fig:demo}
\end{figure}

However, the theoretical bound presented in \cite{fang2019open} is not adaptable to flexible classifiers (\textit{i.e.}, deep neural networks (DNNs)). In Fig.~\ref{fig:demo}, we show that if the classifier is a DNN, the accuracy (OS in Fig. \ref{fig:demo} (b)) in the target domain will drop significantly (yellow line in Fig. \ref{fig:demo} (b)) when minimizing the empirical estimates of the upper bound. This phenomenon confirms that we cannot simply combine the existing theoretical bound and deep algorithms to address the UOSDA problem. 
% According to the upper bound and the definition of open set difference, if empirical open set difference converges to a negative value when minimizing the upper bound, the most target data will be recognized as unknown data. 

To reveal the nature of this phenomenon, we investigate that the lower bound of the distributional discrepancy is the negative value of open set difference.
% Namely, in theory, open set difference should be greater than a negative value that closes to $0$. 
Since DNNs are very flexible and the empirical open set difference can be a negative value, empirical open set difference will be quickly minimized to a very negative value (yellow line in Fig. \ref{fig:demo} (a)).
Based on the lower bound of the distributional discrepancy, if the empirical open set difference is a very small negative number, the distributional discrepancy is greater than a very large positive number. Consequently, we fail to align the distributions of the two domains, resulting in a very low accuracy on the target domain (yellow line in Fig.~\ref{fig:demo} (b)).
% If the converged empirical open set difference is very negative, then its first term closes to $0$ and its second term is a large positive number. Namely, the most target data are recognized as unknown data. As a result, accuracy on the target domain is very low (yellow line in Fig. \ref{fig:demo} (b)).
% Obviously, there is a gap between existing UOSDA upper bound (\textit{i.e.}, Eq. \eqref{eq:osd}) and flexible classifier (\textit{i.e.}, DNNs). Therefore a UOSDA theoretical bound that adapts to DNNs is imperative.

% Why existing bound cannot be used with deep algorithm (overfitting)
% \hl{(YOU need to explain the meaning of $\epsilon $ in this paragraph)} 
In this paper, we propose a new upper bound of target-domain risk for UOSDA (Eq. \eqref{eq: new_ub}), which includes four terms: source-domain risk, \emph{$\epsilon$-open set difference} ($\Delta_\epsilon $), conditional distributional discrepancy between domains, and a constant. $\epsilon $ is the lower bound of open set difference and we construct a new risk estimator $\Delta_\epsilon $ that limits the descent of the open set difference by $\epsilon $. $\Delta_\epsilon $ can ensure the promptly prevention of the lower bound of the distributional discrepancy between two domains from significantly increasing. Fig. \ref{fig:demo} shows that minimizing the empirical estimates of the new upper bound achieves higher accuracy (green line in Fig. \ref{fig:demo}(b)).

% Introduce our method in detail (includes conditional)
Then, we propose a new principle-guided \emph{deep} UOSDA method that trains DNNs via minimizing empirical estimates of the new upper bound. The network structure is shown in Fig. \ref{fig:network}. We employ a generator (${\bm G}$) to extract the feature of input data, a classifier (${\bm C}$) to classify input data, and a domain discriminator (${\bm D}$) to assist distribution alignment. The overall object function consists of source classification loss, binary adversarial loss, domain adversarial loss, and empirical $\Delta_\epsilon $. Specifically, the source classification loss and empirical $\Delta_\epsilon $ are minimized by gradient descent, and a gradient reverse layer is adopted for adversarial losses.

To effectively align distributions between data with known classes, we propose a novel open-set conditional adversarial training strategy based on the tensor product between the feature
representation and the label prediction to capture the multimodal structure of distribution. According to \cite{song2009hilbert, long2018conditional}, it is significant to capture the multimodal structures of distributions using cross-covariance dependency between the features and classes. However, existing deep UOSDA methods align distributions by either the binary adversarial net \cite{saito2018open, feng2019attract} or the multi-binary classifier \cite{liu2019separate}, which is not adequate for distributions with multimodal structure. Furthermore, this novel training strategy also pushes unknown target data away from data with known classes via ${\bm D}$. As shown in Fig. \ref{fig:demo} (b), the novel distribution alignment strategy can further boost the performance of the classifier.

To validate the efficacy of the proposed method, we conduct extensive experiments on several standard benchmark datasets containing $41$ transfer tasks. Compared to existing shallow and deep UOSDA methods, our method shows state-of-the-art performance on digit recognition (\emph{MNIST}, \emph{SVHN}, \emph{USPS}), object recognition (\emph{Office-31}, \emph{Office-Home}) and face recognition (\emph{PIE}). The main contributions of this paper are: 
\begin{itemize}
% \item[1)] We are the first to bridge theoretical bound and deep algorithms for UOSDA problem.
\item[1)] A new theoretical bound of target-domain risk for UOSDA is proposed. It is essential since the existing bound does not apply to flexible classifiers (\textit{i.e.}, DNNs). Thus this work can bridge the gap between the existing theoretical bound and deep algorithms for the UOSDA problem.

% \item[2)] Our UOSDA method is the first deep UOSDA method with the theoretical guarantee.
\item[2)] A UOSDA method based on DNNs is proposed under the guidance of the proposed theoretical bound. The method can better estimate the risk of the classifier on unknown data than existing deep methods with the theoretical guarantee.

\item[3)] A novel open-set conditional adversarial training strategy is proposed to ensure that our method can align the distributions of two domains better than existing UOSDA methods.

% \item[4)] Experiments on $41$ transfer tasks show that our method outperforms the state-of-the-art UOSDA methods. 
\item[4)] Experiments on Digits, Office-31, Office-Home, and PIE show that the accuracy of the OS of our method significantly outperforms all baselines, which shows that our method achieves state-of-the-art performance.
\end{itemize}

% Structure of whole paper
This paper is organized as follows. Section \uppercase\expandafter{\romannumeral2} reviews the works related to UCSDA, open set recognition, and UOSDA. Section \uppercase\expandafter{\romannumeral3} introduces the definitions of notations and our problem. Section \uppercase\expandafter{\romannumeral4} demonstrates the motivation of this paper. Theoretical results and the proposed method are shown in Section \uppercase\expandafter{\romannumeral5}. Experimental results and analyses are provided in Section \uppercase\expandafter{\romannumeral6}. Finally, Section \uppercase\expandafter{\romannumeral7} concludes this paper.


\section{Related Work}
Unsupervised open set domain adaptation is a combination of unsupervised closed set domain adaptation and open set recognition. In this section, we present a systematic review of related studies.

\subsection{Closed Set Domain Adaption}
In \cite{ben2007analysis}, a theoretical bound for UCSDA is given, which indicates that minimizing the source risk and distributional discrepancy is the key to the UCSDA problem. Based on this point, there are two kinds of methods for UCSDA: one is to employ a distributional discrepancy measurer to measure the domain gap \cite{pan2010domain}; the other is the adversarial training strategy \cite{long2018conditional}.

Transfer Component Analysis (TCA) \cite{pan2010domain} utilizes MMD \cite{gretton2012kernel} learning a domain invariant feature by aligning marginal distribution. Meanwhile, Joint Distribution Adaptation (JDA) \cite{long2013transfer} align marginal distribution and conditional distribution simultaneously. In order to simplify the training of a classifier, Easy Transfer Learning (EasyTL) \cite{wang2019easy} exploits the intra-domain information to get a non-parametric feature and the classifier. CORrelation Alignment (CORAL) \cite{sun2016return} aligns second-order statistics of source and target domain to minimize domain divergence. Manifold Embedded Distribution Alignment (MEDA) \cite{wang2018visual} performs a dynamic distribution alignment in a Grassmann manifold subspace.

Meanwhile, deep neural networks have also been introduced into domain adaptation and achieved competitive performance in UCSDA. Deep Adaptation Networks (DAN) \cite{long2015learning} employs the multi-kernel MMD (MK-MMD) to align the feature of 6-8 layers in Alexnet. Deep CORAL Correlation is the extension of shallow method CORAL in deep neural networks. Wasserstein Distance Guided Representation Learning (WDGRL) \cite{shen2018wasserstein} employs the Wasserstein distance to learn an invariant representation in deep neural networks.

Representative adversarial-training-based method are Domain-Adversarial Training of Neural Networks (DANN) \cite{ganin2016domain} and conditional adversarial domain adaptation (CDAN) \cite{long2018conditional}. DANN employs a domain discriminator to recognize which domain data comes from and deceives the domain discriminator by changing features so that an invariant representation can be learned during the adversarial procession. Furthermore, CDAN utilizes the tensor product between feature and classifier prediction to grasp the multimodal information and an entropy condition to control the uncertainty of the classifier.
However, these methods can only cope with the UCSDA problem and are unable to address the UOSDA problem.

\subsection{Open Set Recognition}
This setting allows some unknown classes to be shown in the target domain, but there is no distributional discrepancy between domains. Open Set SVM \cite{jain2014multi} rejects the unknown classes via a fixed threshold. Open Set Nearest Neighbor (OSNN) \cite{junior2017nearest} extends the Nearest Neighbor to recognize unknown classes. Bendale \textit{et al.} \cite{bendale2016towards} introduces a layer named OpenMAX to estimate the probability that an input data is recognized as unknown classes in DNNs. However, these methods do not consider distributional discrepancy. They are also unable to address the UOSDA problem.

\subsection{Open Set Domain Adaptation}
Busto \textit{et al.} \cite{panareda2017open} were the first to propose the setting of UOSDA. They employed a method named Assign-and-Transform-Iterately (ATI) to assign labels to target data using a distance matrix between target data and source class centers and aligned distributions through a mapping matrix. In the setting of this paper, however, the source domain contains some unknown classes to assist the classifier to recognize unknown data. Since obtaining unknown samples of the source domain is expensive and time-consuming, Open Set Backpropagation (OSBP) \cite{saito2018open} assumes a more realistic scenario that the source domain has no unknown classes, which is more challenging. An adversarial network is used to recognize unknown samples and align distribution during backpropagation. 

Based on OSBP, Feng \textit{et al.} \cite{feng2019attract} proposed a method named SCI\_SCM, which utilizes semantic structure among data to align the distribution of known classes and push unknown classes away from known classes. Separate to Adapt (STA) \cite{liu2019separate} utilizes a coarse-to-fine weight mechanism to separate unknown samples from the target domain. In Distribution Alignment with Open Difference (DAOD) \cite{fang2019open}, a theoretical bound is proposed for UOSDA and a risk estimator is used to recognize unknown target data. 

However, existing deep UOSDA methods lack the theoretical guidance and the upper bound in \cite{fang2019open} is not applicable to DNNs, which causes a large distributional discrepancy (details are shown in Section IV). Obviously, for UOSDA, there is a gap between existing theoretical bound and deep algorithms. In this paper, we aim to fill this gap.

\section{Preliminary and Notations}
The definitions of the UOSDA problem and some important concepts are introduced in this section. The notations used in this paper are summarized in Table I.

\subsection{Definitions and Problem Setting}
Important definitions are presented as follows.
\begin{Definition}[{Domain}\cite{fang2019open}]\label{d2} Given a feature space $\mathcal{X}\subset \mathbb{R}^d$ and a label space $\mathcal{Y}$, a \textit{domain} is a joint distribution $P(X, Y)$, where the random variables $X \in \mathcal{X}$, $Y \in \mathcal{Y}$.
\end{Definition}

In Definition 1, $X\in \mathcal{X}$ and  $Y \in \mathcal{Y}$ mean that the spaces $\mathcal{X}$ and $\mathcal{Y}$ contain the image sets of $X$ and $Y$ respectively. In the paper, we name the random variable $X$ as feature vector and the random variable $Y$ as label. Based on this definition, we have:
\begin{Definition}[{Domains for Open Set Domain Adaptation}\cite{fang2019open}]\label{d3}Given a feature space $\mathcal{X}\subset \mathbb{R}^d$ and the label spaces $\mathcal{Y}^s, \mathcal{Y}^t$, the {source} and {target domains} have different joint distributions $P(X^s, Y^s)$ and $P(X^t, Y^t)$, where the random variables $X^s, X^t \in \mathcal{X}$, $Y^s \in \mathcal{Y}^s$, $Y^t \in \mathcal{Y}^t$, and the label space $\mathcal{Y}^s \subset \mathcal{Y}^t$.
\end{Definition}

From the definitions above, we can notice that: 1) This paper focuses on homogeneous situations. Thus $X^s$ and $X^t$ are belong to the same space, and 2) $\mathcal{Y}^t$ contains $\mathcal{Y}^s$. It is \textit{{unknown target classes}} that are the classes from $\mathcal{Y}^t \backslash \mathcal{Y}^s$. It is are the \textit{{known classes}} that are the classes from $\mathcal{Y}^s$. Thus, the UOSDA problem is:

\begin{definition*}[{Unsupervised Open Set Domain Adaptation (UOSDA) \cite{fang2019open}}]
Given labeled samples $\mathcal{S}$ drawn from the joint distribution of the source domain $P(X^s, Y^s)$ i.i.d and unlabeled samples $\mathcal{T}_X$ drawn from the marginal distribution of the target domain $P(X^t)$ i.i.d. {The aim of UOSDA} is to find a target classifier  ${\bm c}^t:\mathcal{X}\rightarrow \mathcal{Y}^{t}$ such that
\newline
\noindent\textit{1)} ${\bm c}^t$ \textit{classifies the known target samples into the correct known classes;}\newline
\noindent\textit{2)} ${\bm c}^t$ \textit{recognizes the unknown target samples as unknown.
}
\end{definition*}

According to the definition of the problem, the target-domain classifier only needs to recognize unknown target data as unknown and classify other target data. It is not necessary to classify unknown target data, and all unknown target data are recognized as the ``unknown class". {In general, we assume that
$\mathcal{Y}^s=\{{\mathbf{y}}_k\}_{k=1}^K,$ $\mathcal{Y}^t=\{{\mathbf{y}}_k\}_{k=1}^{K+1}$, where the label ${\mathbf{y}}_{K+1}$ denotes the unknown class and the label ${\mathbf{y}}_k\in \mathbb{R}^{{(K+1)}\times 1}$ is a one-hot vector. The label ${\mathbf{y}}_k$ denotes the $k$-th class.}

\begin{table*}[t]
\caption{{Notations and their descriptions.}}
\begin{center}
\normalsize
\begin{tabular}{p{2cm}p{6.4cm}|p{2cm}p{6.0cm}}
\hline
Notation & ~~~~~~~~~~Description&Notation & ~~~~~~~~~~Description  \\ \hline
$\mathcal{X}$&  feature space&$P_{X^sY^s},~P_{X^tY^t}$& source, target joint distributions\\
$\mathcal{Y}^s,\mathcal{Y}^t$ & source, target label sets $\{{\mathbf{y}}_c\}_{c=1}^K,\{{\mathbf{y}}_c\}_{c=1}^{K+1}$ &$P_{X^s},~P_{X^t}$&source, target marginal distributions\\
$X^s,X^t$ & random variables on the feature space&$\Delta$& open set difference\\
$Y^s$,~$Y^t$& random variables on the label spaces&$P_{X^t|\mathcal{Y}^s}$&$P(X^t|Y^t\in \mathcal{Y}^s)$
\\
$L^s(\cdot),L^t(\cdot)$&source, target risks&
$L^t_*(\cdot)$&partial risk on known target classes\\
${\mathbf{y}}_c$ & one-hot vector (class $c$)&$L^t_{K+1}(\cdot)$&partial risk on unknown target classes\\ ${\bm G},{\bm C}$ & feature transformation , classifier over ${\hm G}(\mathcal{X})$ &$L^s_{u,K+1},L^t_{u,K+1}$&risks that samples regarded as unknown\\ $\mathcal{H}_{\hm G}$ & hypothesis space, set of classifiers ${\bm C}$&$\pi^t_{K+1}$& class-prior probability for unknown class\\$\mathcal{X}_{\hm G}, \mathbf{x}_{\hm G}$& ${\hm G}(\mathcal{X}),$ sample from ${\hm G}(\mathcal{X})$ &$\widehat{P},\widehat{L}(\cdot)$&empirical distribution, empirical risk\\$d_{{\mathcal{H}_{\hm G}}}^{\ell}(\cdot,\cdot)$ &{$\mathcal{H}$$\Delta$$\mathcal{H}$ distance}& $d_{\Delta_{{\hm C},{\hm G}}}^{\ell}(\cdot,\cdot)$& tensor discrepancy distance\\
\hline
\end{tabular}
\end{center}
%\vspace{-1cm}
\end{table*}
  %Before introducing the details of our method, we provide some basic notations. We define   source labels   $Y_s=[y_1,...,y_{n^s}]\in \mathbb{R}^{(C+1)\times n^s}$, where $C$ is the number of known classes and $y_i^s\in \mathbb{R}^{(C+1)\times1}$ is the label vector such that $y_i^s[l]=1$ if the sample $x_{s_i}$ has label $l$ and $y_i^s[l]=0$, otherwise. Here we use label $C+1$ to represent the unknown target class.
  %Necessary notions and their descriptions are summarized in Table I.
%\vspace{-0.5cm}
\subsection{{Concepts and Notations}}
\vspace{-0.1cm}
It is necessary to introduce some important concepts and notations before demonstrating our main results. Unless otherwise specified, all the following notations are used consistently throughout this paper without further explanations.

\subsubsection{{Notations for distributions}} 
For simplicity, we denote the joint distributions $P(X^s,Y^s)$ and $P(X^t,Y^t)$ by the notations $P_{X^sY^s}$ and $P_{X^tY^t}$ respectively. Similarly, we use $P_{X^s}$ and $P_{X^t}$ denote the marginal distributions $P(X^s)$ and $P(X^t)$ respectively.

$P_{X^t|\mathcal{Y}^s}$ denotes the target conditional distribution for the known classes, while $P_{X^t|\mathbf{y}_{K+1}}$ denotes the target conditional distribution for the unknown classes. $\pi^t_{K+1}=P(Y^t={\mathbf{y}}_{K+1})$ denotes the {class-prior probability for the unknown target classes}. 

Given a feature transformation:
\begin{equation}
\begin{split}
    {\bm G}: \mathcal{X}&\rightarrow {\mathcal{X}}_{\hm G}:={\hm G}({\mathcal{X}})
    \\ \mathbf{x} &\rightarrow \mathbf{x}_{{\hm G}}:={\hm G}(\mathbf{x}),
\end{split}
\end{equation}
the induced distributions related to $P_{X^s}$ and $P_{X^t|\mathcal{Y}^t}$ are
\begin{equation}
\begin{split}
    {\bm G}_{\#}P_{X^s} &:= P( {\bm G}(X^s) );
    \\{\bm G}_{\#}P_{X^t|\mathcal{Y}^s} &:= P({\bm G}(X^t)| Y^t\in \mathcal{Y}^s).
\end{split}
\end{equation}
Lastly, the notation $\widehat{P}$ denotes the corresponding empirical distribution to any distribution $P$. For example, $\widehat{{P}}_{X^sY^s}$ represents the empirical distribution corresponding to ${{P}}_{X^sY^s}$.

\subsubsection{{Risks and Partial Risks}}
In learning theory, risks and partial risks are two important concepts, which are briefly explained below. 

Following the notations in \cite{DBLP:conf/icml/0002LLJ19}, consider a
multi-class classification task with a {\textit{hypothesis space}} $\mathcal{H}_{\bm G}$ of the classifiers
\begin{equation}\label{hyp}
\begin{split}
    ~~~~~~~{\bm {C}}:~\mathcal{X}_{{\bm G}}&\rightarrow \mathcal{Y}^t\\
    {\mathbf{x}}&\rightarrow [C_1({\mathbf{x}}),...,C_{K+1}({\mathbf{x}})]^T.
    \end{split}
\end{equation}

Let
\begin{equation}
\begin{split}
  \ell: \mathbb{R}^{K+1}\times\mathbb{R}^{K+1}&\rightarrow \mathbb{R}_{\geq 0}
  \\ (\mathbf{y}, \tilde{\mathbf{y}} )&\rightarrow \ell(\mathbf{y}, \tilde{\mathbf{y}} ),
  \end{split}
\end{equation}
be the loss function. For convenience, we also require $\ell$ to satisfy the following conditions in Theorem 1:
\\
1. $\ell$ is symmetric and  satisfies triangle inequality;\\
2. $ \ell(\mathbf{y}, \tilde{\mathbf{y}} )=0$ iff $\mathbf{y}= \tilde{\mathbf{y}}$;\\
3. $ \ell(\mathbf{y}, \tilde{\mathbf{y}} )\equiv 1$ if $\mathbf{y}\neq \tilde{\mathbf{y}}$ and $\mathbf{y}, \tilde{\mathbf{y}}$ are one-hot vectors.

We can check many losses satisfying the above conditions such as $0$-$1$ loss $1_{\mathbf{y}\neq \tilde{\mathbf{y}}}$ and $\ell_2$ loss $\frac{1}{2}\|\mathbf{y}- \tilde{\mathbf{y}}\|^2_2$.

Then the {\textit{risks}} of ${\bm C}\in \mathcal{H}_{{\bm G}}$ w.r.t. $\ell$ under ${{\bm G}}_{\#}{P}_{X^sY^s}$ and ${\bm G}_{\#}P_{X^tY^t}$  are given by 
\begin{equation}\label{kk}
   \begin{split}
      L^s({\bm C}\circ{\bm G}):&=\underset{{({\mathbf{x}},{\mathbf{y}})\sim{P}_{X^sY^s}}}{\mathbb{E}}\ell({\bm C}\circ{\bm G}({\mathbf{x}}),{\mathbf{y}}),\\L^t({\bm C}\circ{\bm G}):&=\underset{{({\mathbf{x}},{\mathbf{y}})\sim{P}_{X^tY^t}}}{\mathbb{E}}\ell({\bm C}\circ{\bm G}({\mathbf{x}}),{\mathbf{y}}).
  \end{split}
\end{equation}

The {\textit{partial risk}} of ${\bm C}\in \mathcal{H}_{\bm G}$ for the known target classes is
\begin{equation}\label{k1}
\begin{split}
L_*^t({\bm C}\circ{\bm G}):&=\frac{1}{1-\pi^t_{K+1}}\int_{\mathcal{X}\times \mathcal{Y}^s}\ell({\bm C}\circ{\bm G}({\mathbf{x}}),{\mathbf{y}}){\rm d}P_{X^tY^t}
\end{split}
\end{equation}
and the {\textit{partial risk}} of   ${\bm C}\in \mathcal{H}_{\bm G}$ for the unknown target classes is
\begin{equation}\label{-k}
\begin{split}
   {L}_{K+1}^t( {\bm C}\circ {\bm G}):&=\underset{{{\mathbf{x}}\sim P_{X^t|{\mathbf{y}}_{K+1}}}}{\mathbb{E}} \ell( {\bm C}\circ {\bm G}({\mathbf{x}}),{\mathbf{y}}_{K+1}).
    \end{split}
\end{equation}

Lastly, we denote  
\begin{equation}\label{10000}
\begin{split}
&L^s_{u,K+1}( {\bm C}\circ {\bm G}):=\underset{{\mathbf{x}}\sim P_{X^s}}{\mathbb{E}}\ell( {\bm C}\circ {\bm G}({\mathbf{x}}),{\mathbf{y}}_{K+1}),\\&
L^t_{u,K+1}( {\bm C}\circ {\bm G}):=\underset{{{\mathbf{x}}\sim P_{X^t}}}{\mathbb{E}}\ell( {\bm C}\circ {\bm G}({\mathbf{x}}),{\mathbf{y}}_{K+1})
\end{split}
\end{equation}
as the {\textit{risks}} that the samples are regarded as the unknown classes. 

Given a risk $L( {\bm C}\circ {\bm G})$, it is convenient to use notation $\widehat{L}( {\bm C}\circ {\bm G})$ as the empirical risk that corresponds to $L( {\bm C}\circ {\bm G})$.
\subsubsection{{Discrepancy Distance}}
How to measure the difference between domains plays a critical role in domain adaptation. To achieve this, a famous distribution distance has been proposed as the measures of the distribution difference. 

\begin{Definition}[{Distributional Discrepancy} \cite{DBLP:conf/colt/MansourMR09}]\label{d0}Given a hypothesis space $\mathcal{H}_{\hm G}$ containing a set of functions defined in a feature space $\mathcal{X}_{\hm G}$. Let $\ell$ be a  loss function, and $P_1,P_2$ be distributions on space $\mathcal{X}_{\hm G}$. The {{$\mathcal{H}$$\Delta$$\mathcal{H}$}~distance} $d_{{\mathcal{H}_{\hm G}}}^{\ell}(P_1,P_2)$ between distributions $P_1$ and $P_2$ over $\mathcal{X}_G$ is
\begin{equation*}
    \underset{{\bm C},{\bm C}^*\in \mathcal{H}_{\hm G}}{\sup}\Big|\underset{{{\mathbf{x}}\sim P_1}}{\mathbb{E}}\ell({\bm C}({\mathbf{x}}),{\bm C}^*({\mathbf{x}}))-\underset{{\mathbf{x}}\sim P_2}{\mathbb{E}}\ell({\bm C}({\mathbf{x}}),{\bm C}^*({\mathbf{x}}))\Big|.
\end{equation*}
\end{Definition}

In this paper, we have used a tighter distance named tensor discrepancy distance, which is firstly proposed by \cite{long2018conditional}. The tensor discrepancy distance can future extract the multimodal structure of distributions to make sure  the  knowledge  related  to  learned  classifier  and  pseudo labels can be utilized during the distribution aligning process.

We consider the following tensor mapping:
\begin{equation}
\begin{split}
    {\otimes_{{\hm C}}}: {\mathcal{X}}_G&\rightarrow {\mathcal{X}}_{\hm G} \otimes \mathcal{Y}^t
    \\ \mathbf{x}_{\hm G}&\rightarrow \mathbf{x}_{\hm G}\otimes {\hm C}(\mathbf{x}_{\hm G}).
    \end{split}
\end{equation}

Then we induce two importance distributions:
\begin{equation}
\begin{split}
    {\otimes_{{\hm C}}}_{\#}P_{X^s} &:= P( {\otimes_{{\hm C}}}({\hm G}(X^s)) );
    \\{\otimes_{{\hm C}}}_{\#}P_{X^t|\mathcal{Y}^s}&:= P({\otimes_{{\hm C}}}({\hm G}(X^t)) | Y^t\in \mathcal{Y}^s).
\end{split}
\end{equation}

Using $\mathcal{H}_{\hm G}\subset\{\overline{{\hm C}}: {\mathcal{X}}_{\hm G}\rightarrow  \mathcal{Y}^t\}$, we reconstruct a new hypothetical set:
\begin{equation}
\begin{split}
    &~~~~\Delta_{{\hm C},{\hm G}}:= \{\delta_{\overline{{\hm C}}}:{\mathcal{X}}_{\hm G} \otimes \mathcal{Y}^t\rightarrow \mathbb{R}: \overline{{\hm C}} \in \mathcal{H}_{\hm G} \},
    \end{split}
\end{equation}
where $\delta_{\overline{{\hm C}}}(\mathbf{x}_{\hm G}\otimes \mathbf{y})=|\otimes_{\hm C}(\mathbf{x}_{\hm G})-\otimes_{\overline{{\hm C}}}(\mathbf{x}_{\hm G})|$.
Then the distance between ${\otimes_{{\hm C}}}_{\#}P_{X^s}$ and ${\otimes_{{\hm C}}}_{\#}P_{X^t|\mathcal{Y}^s}$ is:
\begin{equation}
\begin{split}
    &~~~d^{\ell}_{\Delta_{{\hm C},{\hm G}}}({\otimes_{{\hm C}}}_{\#}P_{X^s},{\otimes_{{\hm C}}}_{\#}P_{X^t|\mathcal{Y}^s})\\&=\sup_{\delta\in \Delta_{{\hm C},{\hm G}}}\Big| \underset{{{\mathbf{z}}\sim \otimes_{{\hm C}}}_{\#}P_{X^s}}{\mathbb{E}}{{\rm sgn}\circ \delta(\mathbf{z})}-\underset{{{\mathbf{z}}\sim \otimes_{{\hm C}}}_{\#}P_{X^t|\mathcal{Y}^s}}{\mathbb{E}}{{\rm sgn}\circ\delta(\mathbf{z})}\Big|,
    \end{split}
\end{equation}
where ${\rm sgn}$ is the sign function.

It is easy to prove that under the conditions (1)-(3) for loss $\ell$ and for any ${{\hm C}}\in \mathcal{H}_{\hm G}$, we have
\begin{equation}\label{eq13}
\begin{split}
   & d_{\Delta_{{\hm C},{\hm G}}}^{\ell}({\otimes_{{\hm C}}}_{\#}P_{X^s},{\otimes_{{\hm C}}}_{\#}P_{X^t|\mathcal{Y}^s})\leq
   d^{\ell}_{{\mathcal{H}_{\hm G}}}({{\hm G}}_{\#}P_{X^s},{{\hm G}}_{\#}P_{X^t|\mathcal{Y}^s}).
   \end{split}
\end{equation}

\subsubsection{Existing Theoretical Bound} Zhen \textit{et al.} \cite{fang2019open} firstly proposed a theoretical bound for UOSDA:
\begin{equation}\label{eq:osd}
\begin{split}
 ~~\frac{L^t({\hm C}\circ {\hm G})}{1-\pi_{K+1}^t}&\leq   \overbrace{L^s({\hm C}\circ {\hm G})}^{\text{{Source Risk}}} +{\overbrace{2d_{{\mathcal{H}_{\hm G}}}^{\ell}({{\hm G}}_{\#}P_{X^s},{{\hm G}}_{\#}P_{X^t|\mathcal{Y}^s})}^{\text{{ distributional discrepancy}}}}+\Lambda\\&+\underbrace{ \frac{L^t_{u,K+1}({\hm C}\circ {\hm G})}{1-\pi_{K+1}^t}
 -L^s_{u,K+1}({\hm C}\circ {\hm G})}_{\text{{Open Set Difference}}~\Delta}.
\end{split}
\end{equation}
There are four main terms: source risk, distributional discrepancy, a constant $\Lambda$ and open set difference. The fourth term, open set difference, is designed  to estimate the risk of classifier on unknown data.
%Note that the open set difference should not be smaller than 
%$
%-d_{{\mathcal{H}_{\hm G}}}^{\ell}({{\hm G}}_{\#}P_{X^s},{{\hm G}}_{\#}P_{X^t|\mathcal{Y}^s})
%$.

\section{Motivation}
In UOSDA, the target-domain classifier aims to accurately recognize unknown target data and classify the other target data. Since the knowledge about unknown classes is missing, the classifier is likely to be confused about the boundary between known and unknown target data. Thus, recognizing unknown target data plays a critical role in addressing the UOSDA problem.

In order to obtain an effective target-domain classifier, Zhen \textit{et al.} \cite{fang2019open} have proven an upper (Eq. \eqref{eq:osd}) bound for UOSDA and proposed a \emph{shallow} method based on the bound. It consists of four terms: source-domain risk, distributional discrepancy, \emph{open set difference} ($\Delta$), and a constant. Particularly, open set difference, as an important term, is leveraged to estimate the risk of the classifier on unknown target data.

In order to verify whether open set difference works in DNNs, we introduced open set difference into DNNs and conducted a group of experiments on the task \textit{Ar} $\rightarrow$ \textit{Cl} in \textit{Office-Home}. The classifier consists of backbone (ResNet50), generator (two linear layers), and classifier (one linear layer). It is evident that the classifier is very flexible. As shown in Fig. \ref{fig:demo}, the empirical open set difference converges to a negative value (refer to the yellow line in Fig. \ref{fig:demo}(a)) and the accuracy of {OS}, average accuracy among all classes that include unknown classes (Eq. \eqref{eq:OS}), significantly decreases when empirical open set difference converges to a negative value.

To reveal the nature of this phenomenon, first we investigate the distributional discrepancy and discover that the distributional discrepancy has a lower bound. Specifically, the  distributional discrepancy is greater than the negative value of open set difference (Eq. (\ref{lowerbound})). Based  on  the  lower  bound, if the value of the open set difference is a large negative number, then the distributional discrepancy is greater than a large positive number. Hence, we may fail to align the distributional discrepancy.  In fact, experiments have shown that the empirical open set difference may converge to a large negative value if we introduce the open set difference into DNNs.

Clearly, there is a gap between existing theoretical bound and DNNs. In order to bridge theoretical bound and deep algorithms, in this paper, we propose a new practical upper bound (Eq. \eqref{eq: new_ub}) for UOSDA that applies to DNNs. The term, $\epsilon$-open set difference, in the new bound can effectively overcome the defect of open set difference. As shown in Fig. \ref{fig:demo}, $\epsilon$-open set difference guarantees that the risk of the classifier on unknown data is always greater than the lower bound of open set difference by $\epsilon $ (refer to the green line in Fig. \ref{fig:demo}(a)). Furthermore, the $\epsilon$-open set difference significantly outperforms the open set difference (refer to the green line in Fig. \ref{fig:demo}(b)). 

To sum up, existing upper bound is not compatible with DNNs. That is why we propose a new upper bound that contains an amended risk estimator, $\epsilon$-open set difference ($\Delta_\epsilon $). Details of the new upper bound and $\Delta_\epsilon $ are shown in Section V.

\section{The Proposed Method}
In this section, we firstly propose a theoretical bound that applies to DNNs for UOSDA. Under the guidance of the bound, we then propose a UOSDA method based on DNNs.

\begin{table}[t]\label{table2}
\caption{Notations and their descriptions.}
% \begin{center}
\normalsize
\begin{tabular}{m{1.1cm}|m{6.9cm}}
\hline
Notation & ~~~~~~~~~~Description\\ \hline
$\ell_{ce}, \ell_{mse}$ & cross entropy, mean square error loss function\\ 
\hline
$\mathcal{T}_u^*$ & set of predicted unknown target data with high confidence \\
\hline
$\mathcal{T}_K^{*}$ & set of predicted known target data with high confidence\\
\hline
$n^{s}$ & number of source data \\ \hline
$n^{t}$ & number of target data \\ \hline
$n^*_{K}$ & number of $\mathcal{T}_K^{*}$ \\ \hline
$n_u^{*}$ & number of $\mathcal{T}_u$\\ \hline
$\mathbf{x}^s_i$ & source data \\ \hline
$\mathbf{x}^t_i$ & target data \\
\hline
\end{tabular}
% \end{center}
\vspace{-1em}
\end{table}

\subsection{Theoretical Results}
\subsubsection{An Analysis for Open Set Difference}
Eq. (\ref{eq:eosd}) is the open set difference: 
\begin{equation}\label{eq:eosd}
{\Delta} = \frac{{L}^t_{u,K+1}({\hm C}\circ {\hm G})}{1-\pi_{K+1}^t}-{L}^s_{u,K+1}({\hm C}\circ {\hm G})
\end{equation}
where ${L}^t_{u,K+1}({\hm C}\circ {\hm G})$ and  ${L}^s_{u,K+1}({\hm C}\circ {\hm G})$ are defined in Eq. (\ref{10000}). The positive term ${L}^t_{u,K+1}({\hm C}\circ {\hm G})$ is used to recognize unknown data and the negative term  ${L}^s_{u,K+1}({\hm C}\circ {\hm G})$ is designed to prevent known data from being classified as unknown classes. By combining these two terms, the classifier can recognize unknown target samples. 
%When there is no distributional discrepancy between domains, open set difference should be greater than or equal to 0, but the empirical open set difference, in reality, is likely to less than 0, which has been verified in \cite{kiryo2017positive}. When distributional discrepancy exists between domains, open set difference and empirical open set difference are both likely to less than 0 in UOSDA due to distributional discrepancy. 
According to \cite{fang2019open}, the open set difference $\Delta$ satisfies the following inequality:
\begin{equation}
\label{eq:osd_neq}
\begin{split}
\Delta&=\frac{L^t_{u,K+1}({\hm C}\circ {\hm G})}{(1-\pi_{K+1}^t)}-L^s_{u,K+1}({\hm C}\circ {\hm G}) \\& \geq \frac{\pi_{K+1}^t}{(1-\pi_{K+1}^t)}L^t_{K+1}({\hm C}\circ {\hm G})-d^{\ell}_{{\mathcal{H}_{\hm G}}}({{\hm G}}_{\#}P_{X^s},{{\hm G}}_{\#}P_{X^t|\mathcal{Y}^s}).
\end{split}
\end{equation}
The proof of Eq. \eqref{eq:osd_neq} can be found in  Appendix A. proposition 1. Note that
\begin{equation}
    \frac{\pi_{K+1}^t}{(1-\pi_{K+1}^t)}L^t_{K+1}({\hm C}\circ {\hm G})\geq 0,
\end{equation}
hence, the distributional discrepancy  is greater than the negative open set difference:
\begin{equation}
\label{lowerbound}
    d^{\ell}_{{\mathcal{H}_{\hm G}}}({{\hm G}}_{\#}P_{X^s},{{\hm G}}_{\#}P_{X^t|\mathcal{Y}^s})\geq -\Delta.
\end{equation}
Theoretically, we hope that the optimized open set difference should not be a large negative value. Otherwise, it is impossible to eliminate the distributional discrepancy. However, in fact, the empirical open set difference $\widehat{\Delta}$ may converge to a large negative value (see Fig. \ref{fig:demo}). This results in that the distributional discrepancy may still be large.

\subsubsection{$\mathbf{\epsilon}$-{Open Set Difference}}
Based on the analyses above, we try to correct the open set difference to avoid the problem mentioned above. According to Eq. \eqref{lowerbound}, the open set difference is lower bounded. We denoted the lower bound of the open set difference by $\epsilon $. An potentiality  is to limit the lower bound of the open set difference by a  small negative constant $-\epsilon $. Hence, we propose an amended risk estimator, $\epsilon$-open set difference ($\Delta_\epsilon $), to overcome the existing defect in the open set difference:
\begin{equation}
\label{eq:beta}
\Delta_\epsilon  = \max\{-\epsilon , \frac{{L}^t_{u,K+1}({\hm C}\circ {\hm G})}{1-\pi_{K+1}^t}-{L}^s_{u,K+1}({\hm C}\circ {\hm G})\}.
\end{equation}
If we optimize the empirical $\epsilon$-open set difference, we can guarantee that  the empirical $\epsilon$-open set difference is always larger than $-\epsilon $. Lastly,  combining Eqs. (12), (13) with Eq. (\ref{eq:beta}), we develop a new theoretical bound for UOSDA.

\begin{theorem}\label{-1000}
Given a feature transformation ${\hm G}: \mathcal{X}\rightarrow \mathcal{X}_{\hm G}$, a  loss function $\ell$ satisfying conditions 1-3 introduced in Section III-B-2), a nonegative constant $\epsilon $ and  a hypothesis $\mathcal{H}_{\hm G}\subset\{{\hm C}:\mathcal{X}_{\hm G}\rightarrow \mathcal{Y}^t\}$ with a mild condition that the constant vector value function $\widetilde{{\hm C}}:={\mathbf{y}}_{C+1}\in \mathcal{H}_{\hm G}$, then for any ${\hm C}\in \mathcal{H}_{\hm G}$,  we have 
\begin{equation}
\label{eq: new_ub}
\begin{split}
 ~~&\frac{L^t({\hm C}\circ {\hm G})}{1-\pi_{K+1}^t}\leq   \overbrace{L^s({\hm C}\circ {\hm G})}^{\text{{Source Risk}}} +{\overbrace{2d_{\Delta_{{\hm C},{\hm G}}}^{\ell}({\otimes_{{\hm C}}}_{\#}P_{X^s},{\otimes_{{\hm C}}}_{\#}P_{X^t|\mathcal{Y}^s})}^{\text{{Tensor distributional discrepancy}}}}\\&+\underbrace{\max\{-\epsilon , \frac{L^t_{u,K+1}({\hm C}\circ {\hm G})}{1-\pi_{K+1}^t}
 -L^s_{u,K+1}({\hm C}\circ {\hm G})\}}_{\text{{$\epsilon$-Open Set Difference}}~\Delta_\epsilon }+\Lambda,
 \end{split}
\end{equation}
 where $L^s({\hm C}\circ {\hm G})$ and $L^t({\hm C}\circ {\hm G})$ are the risks defined in (\ref{kk}), $L^s_{u,K+1}({\hm C}\circ {\hm G})$ and $L^t_{u,K+1}({\hm C}\circ {\hm G})$ are the risks defined in (\ref{10000}), $L^t_{*}({\hm C}\circ {\hm G})$ is the partial risk defined in (\ref{k1}) and $\Lambda=\underset{{\hm C}\in \mathcal{H}_{\hm G}}{\min}  ~L^s({\hm C}\circ {\hm G})+L^t_{*}({\hm C}\circ {\hm G})$ . 
\end{theorem}
\begin{proof}
The proof is given in Appendix A.
\end{proof}
It is notable that the theoretical bound introduced in Theorem 1 has two main differences from the learning bound introduced by \cite{fang2019open}. The first one is the $\epsilon$-open set difference. As mentioned before, $\epsilon$-open set difference is designed to eliminate distributional discrepancy caused by open set difference when the module is based on DNNs. The other difference is that we use the tensor distributional discrepancy to estimate the domain difference. There are two 
advantages for the tensor distributional discrepancy compared with the distributional discrepancy (Definition 3): 1) the tensor distributional discrepancy is tighter than the distributional discrepancy (see Eq. (\ref{eq13})); 2) the tensor distributional discrepancy can extract the multimodal structure of distributions to make sure the knowledge related to the learned classifier and pseudo labels can be utilized during the process of distribution alignment \cite{long2018conditional}.
% \vspace{-0.1cm}

\subsection{Method Description}
\begin{figure*}[t]
    \centering
    \includegraphics[scale=0.68, trim=5 130 0 130, clip]{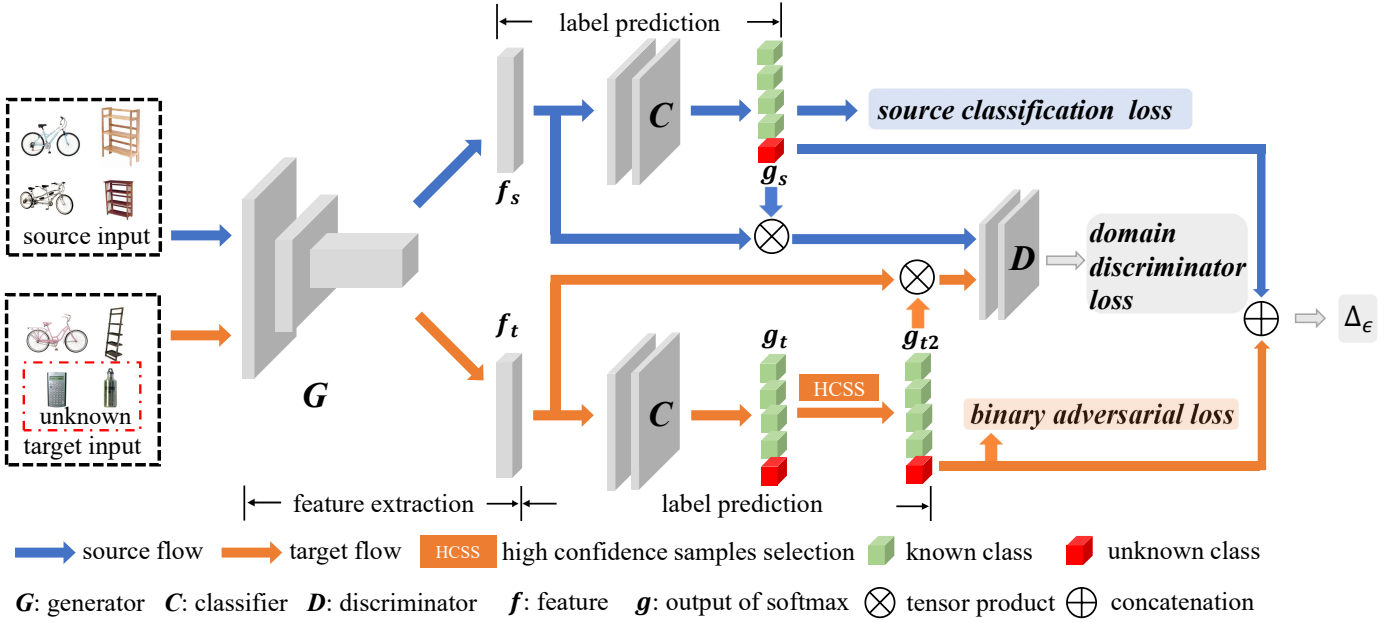}
    \caption{Framework of the proposed method. The generator (${\bm G}$) aims to extract the feature ($f$) of input data and feed it to the classifier (${\bm C}$) to predict its label ($\widehat{y}$). This whole framework consists of three parts. 1) Binary adversarial domain adaptation, which is made of source classification loss and binary adversarial loss. Classifier can find a rough boundary between known data and unknown data. 2) $\epsilon$-open set difference ($\Delta_\epsilon $). We proposed the amended risk estimator to more properly estimate the risk of the classifier on unknown data. 3) Conditional adversarial domain adaptation, which aims to capture multimodal structure of distribution for distribution alignment. In summary, our method can achieve better performance by accurately estimating risk on unknown target data and aligning distribution more adequately.}
    \label{fig:network}
\end{figure*}

According to Theorem 1, we formally present our method (see Fig. \ref{fig:network}), which consists of three parts. Part 1) Binary adversarial domain adaptation. Following \cite{saito2018open}, we employ a binary adversarial module to find a rough boundary between the class-known data (\emph{known data}) and the class-unknown data (\emph{unknown data}), and thus this module can provide target samples with high confidence for other modules. Part 2) $\epsilon$-open set difference ($\Delta_\epsilon $). The $\Delta_\epsilon $ is leveraged to estimate the risk of the classifier on unknown data such that the classifier can accurately recognize the unknown target data. Part 3) Conditional adversarial domain adaptation. Existing deep UOSDA methods ignore the importance of the multimodal structure of distribution while aligning distributions for known classes. According to the tensor distributional discrepancy, we design a novel open set conditional adversarial strategy to align distributions for known classes. Notations used in this section are summarized in Table \uppercase\expandafter{\romannumeral2}.
\subsubsection{Binary adversarial domain adaptation (BADA)}
% Separating unknown data from the target domain is the first step in UOSDA. The feature generator consists of a stack of convolution neural networks (CNN) and fully-connect layers, which attempts to grasp the feature representation of input data. The classifier consists of one fully-connect layer, which is employed to predict labels of samples. 
According to our theoretical bound, the first term is source risk. For the source domain, the label is available. We utilize a cross-entropy for the classification of source samples:
\begin{equation}
\label{loss:ls}
    \widehat{L}_{cls}^s = \frac{1}{n^s}\sum_{i=1}^{n^s}\ell_{ce}({\bm C}\circ {\bm G}(\mathbf{x}_i^s), \mathbf{y}_i^s)
\end{equation}
For the target domain, it is imperative to recognize the unknown target data before aligning distribution. Following \cite{saito2018open}, we employ a binary cross-entropy and a gradient reverse layer between generator and classifier to find a boundary between the known data and the unknown data:
\begin{equation}
\begin{split}
\label{loss:badv}
    &\widehat{L}_{badv} \\=& -\frac{1}{2n^t}\sum_{i=1}^{n^t}\log \big(({C}_{K+1}\circ {\bm G}(\mathbf{x}_i^t))\big) \big(1-({C}_{K+1}\circ {\bm G}(\mathbf{x}_i^t))\big),
\end{split}
\end{equation}
where ${C}_{K+1}$ is the $K+1$-th value of hypothesis function ${\hm C}$.
% \begin{equation}
% \label{loss:badv}
%     \widehat{L}_{badv} = \frac{1}{n^t}\sum_{i=1}^{n^t}\ell_{bce}(P(y=K+1|\mathbf{x}_i^t), \frac{1}{2})
% \end{equation}

The minimax game is shown in  Section V-C. During the process of adversarial training, the classifier attempts to minimize $\widehat{L}_{badv}$, but the generator attempts to maximize $\widehat{L}_{badv}$. Therefore, recognizing unknown data is achieved during the process of adversarial training.

However, this module can only find a coarse boundary between the known data and the unknown data, which cannot accurately recognize the unknown target data. Table VI verifies that only binary adversarial domain adaptation cannot achieve satisfactory performance. Therefore, we employ the $\epsilon$-open set difference for recognizing unknown target data more appropriately and the open-set conditional adversarial strategy to further align distribution.

\subsubsection{$\epsilon$-open set difference}
The principle of the $\epsilon$-open set difference ($\Delta_\epsilon$) is adequately demonstrated in Sections IV and V-A. Then we introduce $\Delta_\epsilon$ to recognize unknown target data. According to Eqs. (\ref{eq:beta}), (\ref{loss:eqbeta0}), we can calculate the empirical $\epsilon$-open set difference $\widehat{\Delta}_\epsilon$ by:
\begin{equation}
\label{loss:eqbeta0}
\begin{split}
 &\max\{-\epsilon , \frac{\alpha}{n^t}\sum_{i=1}^{n^t}\ell_{mse}({\bm C}\circ {\bm G}(\mathbf{x}_i^t), \mathbf{y}_{K+1})\\&-\frac{1}{n^s}\sum_{i=1}^{n^s}\ell_{mse}({\bm C}\circ {\bm G}(\mathbf{x}_i^s), \mathbf{y}_{K+1})\}.
\end{split}
\end{equation}
Without more label information, $\pi_{K+1}^t$ in Eq.(\ref{eq:beta}) is impossible to be evaluated accurately, thus, we introduce a parameter, $\alpha$, to replace it. The analysis of $\alpha$ is discussed in Section VI.

\subsubsection{Conditional adversarial domain adaptation}
Here we utilize the tensor distributional discrepancy to align the distribution between the known classes. Firstly, the empirical representations of ${\otimes_{{\hm C}}}_{\#}\widehat{P}_{X^s}$ and ${\otimes_{{\hm C}}}_{\#}\widehat{P}_{X^t|\mathcal{Y}^s}$ can be written as follows:
\begin{equation}
\begin{split}
    &{\otimes_{{\hm C}}}_{\#}\widehat{P}_{X^s}=\frac{1}{n^s}\sum_{i=1}^{n^s} 
    1_{{\hm G}(\mathbf{x}_i^s)\otimes {\hm C}\circ{\hm G}(\mathbf{x}_i^s)},
\\&  {\otimes_{{\hm C}}}_{\#}\widehat{P}_{X^t|\mathcal{Y}^s}=\frac{1}{|\mathcal{T}_K|}\sum_{\mathbf{x}\in \mathcal{T}_X} 
    1_{{\hm G}(\mathbf{x})\otimes {\hm C}\circ{\hm G}(\mathbf{x})},
\end{split}
\end{equation}
where $\mathcal{T}_K$ is the set of target data from the known classes and  $1_{{\hm G}(\mathbf{x})\otimes {\hm C}\circ{\hm G}(\mathbf{x})}$ is the Dirac measure.

Then, motivated by DANN \cite{ganin2016domain} and CDAN \cite{long2018conditional} , we can reformulate the tensor distributional discrepancy between the known classes as follows:
\begin{equation}
\label{eq:dadv_na}
\begin{split}
    &-\frac{1}{n^{s}}\sum_{i=1}^{n^{s}}\log \big({\bm D}({\bm G}(\mathbf{x}_i^{s})\otimes {\bm C} \circ {\bm G}(\mathbf{x}_i^s))\big)\\&-\frac{1}{|\mathcal{T}_K|}\sum_{\mathbf{x}\in \mathcal{T}_K}\big(1-{\bm D}(\log({\bm G}(\mathbf{x})\otimes {\bm C} \circ {\bm G}(\mathbf{x})))\big),
    \end{split}
\end{equation}
where ${\bm D}$ is the domain discriminator designed to classify domains.   
% According to \cite{song2009hilbert, long2018conditional}, it is significant to capture multimodal structures of distributions using cross-covariance dependency between the features and classes. However, existing deep UOSDA methods align distributions by either binary adversarial net \cite{saito2018open, feng2019attract} or multi-binary classifier \cite{liu2019separate}, which is not adequate to align distributions with multimodal structure and cannot perform excellently.

% In order to capture multimodal structure of distribution, We introduce a tensor product \cite{long2018conditional} between feature representation (${\bm G}(\mathbf{x})$) and label prediction (${\bm C}\circ{\bm G}(\mathbf{x})$).
Since the target data is unlabeled, Eq. \eqref{eq:dadv_na} cannot be directly calculated. Thanks to the pseudo labels provided by BADA, we leverage it to replace the true label. Since these pseudo labels are not completely accurate, we only select the samples with a confidence of 0.9. We then formulate the domain adversarial loss function below.
\begin{equation}
\label{loss:dadv}
\begin{split}
    &\widehat{L}_{dadv}=-\frac{1}{n^{s}}\sum_{i=1}^{n^{s}}\log \big({\bm D}({\bm G}(\mathbf{x}_i^{s})\otimes {\bm C} \circ {\bm G}(\mathbf{x}_i^s)) \big)\\&-\frac{1}{n^{*}_{K}}\sum_{\mathbf{x}\in \mathcal{T}^*_K}\big(1-{\bm D}(\log({\bm G}(\mathbf{x})\otimes {\bm C} \circ {\bm G}(\mathbf{x})))\big),
    \end{split}
\end{equation}
where $\mathcal{T}^*_K$ denotes the set of samples from known classes with high confidence in the target domain, and $n^*_K=|\mathcal{T}^*_K|$.

Domain adversary loss aims to minimize over ${\bm D}$ and maximize over ${\bm G}$. The gradient reverse layer between ${\bm G}$ and ${\bm D}$ results in ${\bm D}$ becoming confused about the source data and the target data. The minimax game is shown in Section V-C. The classifier aims to identify what input data belongs to which domain, but the generator aims to deceive the classifier by changing the features of the input data. Distribution alignment can be achieved during this process.

% \vspace{-0.05cm}
Furthermore, the unknown data may distract distribution alignment of the known data. Thus the unknown data should be pushed away from known data to prevent them from affecting distribution alignment. We construct the loss function below. It is worth noting that there is no gradient reverse between ${\bm D}$ and ${\bm G}$ during the process of backpropagation.
% which enable domain discriminator find a boundary between known data and unknown data.
\begin{equation}
\label{loss:d}
\begin{split}
\widehat{L}_{d}=&-\frac{1}{n^{s}}\sum_{i=1}^{n^{s}}\log \big({\bm D}({\bm G}(\mathbf{x}_i^{s})\otimes {\bm C} \circ {\bm G}(\mathbf{x}_i^s))\big)\\&-\frac{1}{n_{u}^{*}}\sum_{\mathbf{x}\in\mathcal{T}_u^*}\big(1-{\bm D}(\log({\bm G}(\mathbf{x})\otimes {\bm C} \circ {\bm G}(\mathbf{x})))\big),
\end{split}
\end{equation}
where $\mathcal{T}_u^*$ is the unknown target samples with high confidence and $n_u^*=|\mathcal{T}_u^*|$.

In this subsection, we construct a domain discriminator (${\bm D}$) to align the distributions for the known data by a tensor product, which can capture the multimodal structure of distribution. Furthermore, we construct a loss function to push the unknown data away from the known data to prevent the unknown data affecting distribution alignment.

\subsection{Training Procedure}
Combining Eqs. \eqref{loss:ls}, \eqref{loss:badv}, \eqref{loss:eqbeta0}, \eqref{loss:dadv} and \eqref{loss:d}, We solve UOSDA problem by the following minimax game:
\begin{equation}
\begin{split}
&\min_{\bm G} \widehat{L}_{cls}^s - \widehat{L}_{badv} + \widehat{\Delta}_\epsilon-\widehat{L}_{dadv} + \widehat{L}_d, \\
&\min_{\bm C} \widehat{L}_{cls}^s + \widehat{L}_{badv}  + \widehat{\Delta}_\epsilon ,\\
&\min_{\bm D} \widehat{L}_{dadv} + \widehat{L}_d.
\end{split}
\end{equation}
We introduce the gradient reverse layer for adversary learning. The whole training procedure is shown in Algorithm 1. Firstly, we initialize the parameters of the generator (${\bm G}$),  the classifier (${\bm C}$) and the domain discriminator (${\bm D}$) (line 1). In each epoch, we divide data into multi minibatches (line 4-5). Then we calculate source risk ($\widehat{L}_{cls}^s$), binary adversarial loss ($\widehat{L}_{badv}$) and $\Delta_\epsilon$ according to Eqs. \eqref{loss:ls}, \eqref{loss:badv}, \eqref{loss:eqbeta0} (line 6-7). After selecting target samples with high confidence ($\geq 0.9$) (line 8), we calculate $\widehat{L}_{dadv}$ and $\widehat{L}_d$ according to Eqs. \eqref{loss:dadv} and \eqref{loss:d} (line 9). Finally, parameters are updated Via the SGD optimizer (line 10). 

With the proposed method, in binary adversarial domain adaptation ($\widehat{L}_{cls}^s$, $\widehat{L}_{badv}$), a coarse boundary between known data and unknown data can be found. Furthermore, $\epsilon$-open set difference ($\widehat{\Delta}_\epsilon$) can adequately estimate the risk of the classifier on unknown data, which is effective for the classifier to accurately recognize unknown target data. Then, we further align distributions of known data ($\widehat{L}_{dadv}$) and push unknown data away from known data ($\widehat{L}_d$) using a domain discriminator. Finally, combining these three modules, we can adequately solve the UOSDA problem.

\begin{algorithm}[tb]
\small
\caption{Training procedure of our method}
\label{alg:training}
\textbf{Input}: source samples $\{\mathbf{x}_{i}^s, \mathbf{y}_{i}^s\}_{i=1}^{n^s}$, target samples $\{\mathbf{x}_{i}^{t}\}_{i=1}^{n^t}$.\\
\textbf{Parameter}: learning rate $\gamma$, batch size $m$, the number of iteration $T$, network parameters $\theta_{{\bm G}}$, $\theta_{{\bm C}}$, $\theta_{{\bm D}}$.\\
\textbf{Output}: predicted target label $\widehat{\mathbf{y}}_t$.\\
\begin{algorithmic}[1]
\STATE Initialize $\theta_{{\bm G}}$, $\theta_{{\bm C}}$, $\theta_{{\bm D}}$
% \STATE pretrain G and C on $\mathbf{X}_s$, update $\theta_{{\bm G}}$, $\theta_{{\bm C}}$
\STATE $t$=0
\WHILE{$t<T$}
\STATE sample source minibatch \{$(\mathbf{x}_{i_1}^{s}, \mathbf{y}_{i_1}^{s}),\dots,(\mathbf{x}_{i_m}^s, \mathbf{y}_{i_m}^s)$\}.
\STATE sample target minibatch \{${\mathbf{x}_{i_1}^t},\dots,{\mathbf{x}_{i_m}^t}$\}.
\STATE calculate $\widehat{L}_s$, $\widehat{L}_{badv}$ according to Eqs. (\ref{loss:ls}) and (\ref{loss:badv}).
\STATE calculate $\widehat{\Delta}_\epsilon $ according to Eq. (\ref{loss:eqbeta0}).
% \IF{$t > T_c$}
\STATE select high confidence target samples according to the \\output of softmax $g_t$.
\STATE calculate $L_{dadv}, L_d$ according to Eqs. (\ref{loss:dadv}) and (\ref{loss:d}) by leveraging high confidence target samples.
% \ENDIF
% \STATE back high confidence target samples into C to train again and calculate $L_{t2}$
\STATE update parameter: 
$\theta_{{\bm G}} = \theta_{{\bm G}} - \gamma\bigtriangledown_{\theta_{{\bm G}}}(\widehat{L}_{cls}^s - \widehat{L}_{adv} + \widehat{\Delta}_\epsilon  - \widehat{L}_{dadv} + \widehat{L}_d)$\\
$\theta_{{\bm C}} = \theta_{{\bm C}} - \gamma\bigtriangledown_{\theta_{{\bm C}}}(\widehat{L}_{cls}^s + \widehat{L}_{adv}  + \widehat{\Delta}_\epsilon )$\\ 
% \IF{$t > T_c$}
% \STATE
$\theta_{{\bm D}}=\theta_{{\bm D}}-\gamma\bigtriangledown_{\theta_{{\bm D}}}(\widehat{L}_{dadv} + \widehat{L}_d)$.
% \ENDIF
\STATE $t=t+1$
\ENDWHILE
\end{algorithmic}
\end{algorithm}

\section{Experiments And Evaluations}
In this section, we conducted extensive experiments on $6$ standard benchmark datasets (including $41$ transfer tasks) to demonstrate the effectiveness of our method. Several state-of-the-art UOSDA methods such ATI-$\lambda$ \cite{panareda2017open}, OSBP \cite{saito2018open}, SCI\_SCM \cite{feng2019attract}, STA \cite{liu2019separate} and DAOD \cite{fang2019open} are employed as our baselines.

\subsection{Datasets}
\textbf{Digits} contains three digit datasets: \textit{MNIST} (\textit{M}) \cite{lecun1998gradient}, \textit{SVHN} (\textit{S}) \cite{netzer2011reading}, \textit{USPS} (\textit{U}) \cite{hull1994database}. We construct three open set domain adaptation tasks as previous works \cite{saito2018open}: \textit{S} $\rightarrow$ \textit{M}, \textit{M} $\rightarrow$ \textit{U} and \textit{U} $\rightarrow$ \textit{M}. Following the protocol of \cite{saito2018open}, we select classes $0$-$4$ as the known classes and classes $5$-$9$ as the unknown classes of the target domain.

\textbf{Office-31} \cite{saenko2010adapting} is an object recognition dataset with $4,110$ imges, which consists of three domains with slight discrepancy: \textit{amazon} (\textit{A}), \textit{dslr} (\textit{D}) and \textit{webcam} (\textit{W}). Each domain contains $31$ kinds of object. So there are $6$ open set domain adaptation tasks on \textit{Office-31}: \textit{A} $\rightarrow$ \textit{D}, \textit{A} $\rightarrow$ \textit{W}, \textit{D} $\rightarrow$ \textit{A}, \textit{D} $\rightarrow$ \textit{W}, \textit{W} $\rightarrow$ \textit{A}, \textit{W} $\rightarrow$ \textit{D}. We follow the open set protocol of \cite{saito2018open}, selecting the first $10$ classes in alphabetical order as the known classes and classes $21$-$31$ as the unknown classes of the target domain.

\textbf{Office-Home} \cite{venkateswara2017deep} is an object recognition dataset with $15,500$ image, which contains four domains with more obvious domain discrepancy than \textit{Office-31}. These domains are \textit{Artistic} (\textit{Ar}), \textit{Clipart} (\textit{Cl}), \textit{Product} (\textit{Pr}), \textit{Real-World} (\textit{Rw}). Each domain contains $65$ kinds of objects. So there are $12$ open set domain adaptation tasks on \textit{Office-Home}: \textit{Ar} $\rightarrow$ \textit{Cl}, \textit{Ar} $\rightarrow$ \textit{Pr}, \textit{Ar} $\rightarrow$ \textit{Rw}, ..., \textit{Rw} $\rightarrow$ \textit{Pr}. Following the standard protocol, we chose the first $25$ classes as the known classes and $26$-$65$ classes as the unknown classes of the target domain.

\textbf{PIE} \cite{Rasouli_2019_ICCV} is a face recognition dataset, containing $41,368$ images of $68$ people with multifarious pose, illumination and expression. following the protocol of \cite{fang2019open}, We performed open set domain adaptation among $5$ out of $13$ poses and selected classes $1$-$20$ as the known classes and classes $21$-$68$ as the unknown classes of the target domain:x \textit{PIE1} (left pose), \textit{PIE2} (upward pose), \textit{PIE3} (downward pose), \textit{PIE4} (frontal pose) and \textit{PIE5} (right pose). We construct $20$ open set domain adaptation tasks, \textit{i.e.}, \textit{PIE1} $\rightarrow$ \textit{PIE2}, \textit{PIE1} $\rightarrow$ \textit{PIE3}, ...,  \textit{PIE5} $\rightarrow$ \textit{PIE4}.

\subsection{Implementation}
\textbf{Network structure}. For the \textit{Digit}, we employ the similar convolution neural network as \cite{shu2018a, saito2018open} for \textit{S} $\rightarrow$ \textit{M} and other tasks, respectively, and train the DNNs from scratch. For \textit{Office-31}, we leverage VGGNet \cite{simonyan2014very} as backbone to extract features of images. We employ two fully-connected layers as the generator and one fully-connected layer as the classifier. For \textit{Office-Home}, We leverage ResNet-$50$ \cite{he2016deep} as backbone to extract features of images. The network structure of the generator and the classifier are the same as \textit{Office-31}. \text{PIE} has provided valid features of all images. Therefore CNN is not necessary, and we adopted a similar generator and classifier as \textit{Office-31}. Details about the network can be found in Appendix B. In the same manner as \cite{saito2018open, feng2019attract}, we do not update the parameters of the backbone during the training process.

\textbf{Parameter setting}. In the proposed method, there are two important parameters: $\alpha$ and $\epsilon $. We set $\epsilon $ as $0$ in all experiments, which is because distributional discrepancy is gradually approaching to $0$ during the process of domain adaptation and $\Delta_\epsilon $ should be greater than or equal to $0$ when distributional discrepancy is $0$. Besides, we set $\alpha$  as $1.25$ for \textit{Office-31}, $1.1$ for \textit{Digit} and \textit{Office-Home}, and $1.0$ for \textit{PIE}. When the distributional discrepancy is relatively large, we advise that $\alpha$ should be smaller for steady training. All experiment results are the accuracy averaged over three independent runs. 
% For the trade-off parameters, $\mu_1$ is set as 1 in \textit{Office-31} and \textit{Digits}, 0.15 in \textit{Office-Home}, and 0.3 in \textit{PIE}. $\mu_2$ is set as 1 in \textit{Office-31}, 0.5 in \textit{Office-Home}, and 0.1 in \textit{PIE} and \textit{Digits}. All experiment results are the accuracy averaged over three independent runs.

\subsection{Baselines}
We compare our method with five UOSDA methods: ATI-{$\lambda$}, {OSBP} \cite{saito2018open}, {SCA\_SCM} \cite{feng2019attract}, {STA} \cite{liu2019separate}, and {DAOD} \cite{fang2019open}. We briefly introduce these baselines in the following.\\
$\bullet$ ~{ATI-$\lambda$} \cite{panareda2017open} employs an integer programming to assign the label for the target domain and a mapping matrix to align distribution.\\
$\bullet$ ~{OSBP} \cite{saito2018open} employs a classifier to align distributions between data (with known classes) in both source and target domains and an adversarial net to reject unknown samples through the probability of samples in the target domain.\\
$\bullet$ {SCA\_SCM} \cite{feng2019attract} aligns the centroids between source and target and pushes unknown samples away from known classes to achieve a good performance.\\ $\bullet$
~{STA} \cite{liu2019separate} utilizes a coarse-to-fine weight mechanism to separate unknown samples from the target domain and achieves distribution alignment simultaneously.
\\ $\bullet$
~{DAOD} \cite{fang2019open} trains a target-domain classifier via minimizing Eq. \eqref{eq:osd}. The term, open set difference, is used to estimate the risk of the classifier on unknown classes.

\subsection{Evaluation Metrics}
Following previous works \cite{panareda2017open, saito2018open, fang2019open}, we employ the two metrics below to evaluate our method. \textbf{OS}: average accuracy among all classes that include unknown classes. \textbf{OS*}: average accuracy among known classes.
\begin{equation}
\begin{split}
\label{eq:OS}
&{\rm Acc(OS^*)} = \frac{1}{K}\sum_{c=1}^{K}\frac{|\mathbf{x}\in\mathcal{T}_c\bigwedge {\bm g}^t(\mathbf{x})=\mathbf{y}_c|}{|\mathcal{T}_c|}\\ 
&{\rm Acc(OS)} = \frac{1}{K+1}\sum_{c=1}^{K+1}\frac{|\mathbf{x}\in\mathcal{T}_c\bigwedge {\bm g}^t(\mathbf{x})=\mathbf{y}_c|}{|\mathcal{T}_c|}\\ 
% &{\rm Acc(UNK) }= \frac{|\mathbf{x}\in\mathbf{X}_{K+1}^t\bigwedge {\bm C}\circ{\bm G}(\mathbf{x})=K+1|}{|\mathbf{X}^t_{K+1}|}\\
\end{split}
\end{equation}
where ${\bm g}^t$ is the target classifier, and $\mathcal{T}_k$ is the set of target samples with label $\mathbf{y}_c$.
% It is worth noting that: When the value of {OS*} is greater than  the value of {OS}, the accuracy in known classes is greater than the accuracy in unknown classes, and vice versa.

\subsection{Results}
\begin{table*}[htbp] %开始一个表格environment，表格的位置是h,here。
%\tiny
% \vspace{-1em}
\centering
\caption{\label{tab:digit} Acc(OS*) and Acc(OS) (\%) on \textit{Digits}} %显示表格的标题
\begin{tabular}{p{1.25cm}p{0.75cm}<{\centering}p{0.75cm}<{\centering}p{0.75cm}<{\centering}p{0.75cm}<{\centering}p{0.75cm}<{\centering}p{0.75cm}<{\centering}p{0.75cm}<{\centering}p{0.75cm}<{\centering}p{0.75cm}<{\centering}p{0.75cm}<{\centering}p{0.75cm}<{\centering}p{0.75cm}<{\centering}} %设置了每一列的宽度，强制转换。
\hline
\multirow{2}{*}{Dataset} &\multicolumn{2}{c}{ATI-$\lambda$}&\multicolumn{2}{c}{OSBP} &\multicolumn{2}{c}{SCA\_SCM} &\multicolumn{2}{c}{STA} &\multicolumn{2}{c}{DAOD}&\multicolumn{2}{c}{OURS} \\  
\cline{2-13}
&OS&OS*&OS&OS*&OS&OS*&OS&OS*&OS&OS*&OS&OS*\\
\hline
\textit{S} ~$\rightarrow$ \textit{M} &67.6&66.5 &63.1&59.1 &68.6&65.5 &76.9&75.4 &-&- &\textbf{82.9}&\textbf{82.6} \\
\textit{M} $\rightarrow$ \textit{U} &86.8&89.6 &92.1&94.9 &91.3&92.0 &{93.0}&\textbf{94.9} &-&- &\textbf{93.4}&94.6 \\
\textit{U} $\rightarrow$ \textit{M} &82.4&81.5 &92.3&91.2 &\textbf{93.1}&\textbf{95.2} &92.2&91.3 &-&- &90.7&92.7 \\
\hline
Average         &78.9&79.2 &82.4&81.7 &84.3&84.2 &87.3&87.2 &-&- &\textbf{89.0}&\textbf{90.0} \\
\hline
\end{tabular}
\vspace{-1em}
\end{table*}

\begin{table*}[htbp] %开始一个表格environment，表格的位置是h,here。
\centering
\caption{\label{tab:office} Acc(OS*) and Acc(OS) (\%) on \textit{Office-31} (VGG-19) and \textit{Office-Home} (Resnet-50).} %显示表格的标题
\begin{tabular}{p{1.25cm}p{0.75cm}<{\centering}p{0.75cm}<{\centering}p{0.75cm}<{\centering}p{0.75cm}<{\centering}p{0.75cm}<{\centering}p{0.75cm}<{\centering}p{0.75cm}<{\centering}p{0.75cm}<{\centering}p{0.75cm}<{\centering}p{0.75cm}<{\centering}p{0.75cm}<{\centering}p{0.75cm}<{\centering}} %设置了每一列的宽度，强制转换。
\hline
\multirow{2}{*}{Dataset} &\multicolumn{2}{c}{ATI-$\lambda$}&\multicolumn{2}{c}{OSBP} &\multicolumn{2}{c}{SCA\_SCM} &\multicolumn{2}{c}{STA} &\multicolumn{2}{c}{DAOD}&\multicolumn{2}{c}{OURS} \\  
\cline{2-13}
&OS&OS*&OS&OS*&OS&OS*&OS&OS*&OS&OS*&OS&OS*\\
\hline
\textit{A} $\rightarrow$ \textit{D} &79.8&86.8 &85.8&85.8 &90.1&92.0 &88.6&92.8 &89.2&91.1  &\textbf{96.0}&\textbf{97.5}  \\
\textit{A} $\rightarrow$ \textit{W} &86.4&93.0 &76.9&76.6 &86.4&87.7 &91.9&\textbf{94.3} &90.5&91.9  &\textbf{92.5}&93.7  \\
\textit{D} $\rightarrow$ \textit{A} &75.0&81.5 &\textbf{89.4}&\textbf{91.5} &81.6&88.4 &73.4&74.3 &75.4&73.6  &85.3&86.0  \\
\textit{D} $\rightarrow$ \textit{W} &91.7&98.6 &96.0&96.6 &97.9&99.8 &96.5&99.5 &\textbf{98.6}&\textbf{100.0} &98.4&\textbf{100.0} \\
\textit{W} $\rightarrow$ \textit{A} &75.8&82.0 &\textbf{83.4}&83.1 &80.3&82.6 &71.3&71.3 &75.6&74.7  &83.2&\textbf{83.9}  \\
\textit{W} $\rightarrow$ \textit{D} &91.5&99.3 &97.1&97.3 &98.2&99.3 &95.4&\textbf{100.0} &98.6&99.3  &\textbf{98.6}&\textbf{100.0} \\
\hline
Average         &83.4&90.2 &88.0&88.5 &89.1&91.6 &86.2&88.7 &88.0&88.4 &\textbf{92.3}&\textbf{93.5} \\
\hline
\hline
\textit{Ar} $\rightarrow$ \textit{Cl} &53.1&54.2  &53.1&53.3  &58.9 &59.9  &57.0&59.3  &55.4&55.3  &\textbf{61.6}&\textbf{62.8} \\
\textit{Ar} $\rightarrow$ \textit{Pr} &68.6&70.4  &68.4&69.2  &73.4 &74.4  &67.2&69.5  &71.8&72.6  &\textbf{76.6}&\textbf{78.3} \\
\textit{Ar} $\rightarrow$ \textit{Rw} &77.3&78.1  &78.0&79.1  &79.2 &80.2  &79.1&81.9  &77.6&78.2  &\textbf{83.2}&\textbf{85.0} \\
\textit{Cl} $\rightarrow$ \textit{Ar} &57.8&59.1  &57.9&58.2  &60.6 &61.5  &59.1&61.3  &59.2&59.1  &\textbf{62.2}&\textbf{62.8} \\
\textit{Cl} $\rightarrow$ \textit{Pr} &66.7&68.3  &\textbf{71.6}&\textbf{72.4}  &67.5 &68.4  &63.4&65.9  &70.1&70.8  &71.0&72.2 \\
\textit{Cl} $\rightarrow$ \textit{Rw} &74.3&75.3  &71.4&72.3  &74.8 &75.8  &72.7&75.5  &77.0&77.8  &\textbf{77.7}&\textbf{79.0} \\
\textit{Pr} $\rightarrow$ \textit{Ar} &61.2&62.6  &59.6&61.0  &63.8 &64.7  &63.8&65.2  &\textbf{65.8}&\textbf{66.7}  &64.6&65.4 \\
\textit{Pr} $\rightarrow$ \textit{Cl} &53.9&54.1  &55.7&56.9  &58.1 &59.0  &56.5&58.6  &59.1&60.0  &\textbf{60.0}&\textbf{60.8} \\
\textit{Pr} $\rightarrow$ \textit{Rw} &79.9&81.1  &82.1&83.9  &77.7 &78.7  &80.1&82.4  &\textbf{82.2}&\textbf{84.1}  &81.5&82.9 \\
\textit{Rw} $\rightarrow$ \textit{Ar} &70.0&70.8  &66.5&68.2  &67.3 &68.2  &69.3&71.3  &70.5&71.3  &\textbf{70.6}&\textbf{71.6} \\
\textit{Rw} $\rightarrow$ \textit{Cl} &55.2&55.4  &57.8&59.2  &55.8 &56.7  &57.5&59.2  &57.8&58.4  &\textbf{58.8}&\textbf{59.6} \\
\textit{Rw} $\rightarrow$ \textit{Pr} &78.3&79.4  &78.6&80.8  &77.7 &78.6  &79.4&82.2  &80.6&81.8  &\textbf{81.3}&\textbf{82.8} \\
\hline
Average &66.4&67.4  &66.7&67.9  &67.9&68.8  &67.1&69.4  &68.9&69.6 &\textbf{70.8}&\textbf{71.9} \\
\hline
\end{tabular}
\vspace{-1em}
\end{table*}

\begin{table*}[htbp] %开始一个表格environment，表格的位置是h,here。
\centering
\caption{\label{tab:pie} Acc(OS*) and Acc(OS) (\%) on \textit{PIE}.} 
\begin{tabular}{p{1.25cm}p{0.75cm}<{\centering}p{0.75cm}<{\centering}p{0.75cm}<{\centering}p{0.75cm}<{\centering}p{0.75cm}<{\centering}p{0.75cm}<{\centering}p{0.75cm}<{\centering}p{0.75cm}<{\centering}p{0.75cm}<{\centering}p{0.75cm}<{\centering}p{0.75cm}<{\centering}p{0.75cm}<{\centering}} %设置了每一列的宽度，强制转换。
\hline
\multirow{2}{*}{Dataset} &\multicolumn{2}{c}{ATI-$\lambda$}&\multicolumn{2}{c}{OSBP} &\multicolumn{2}{c}{SCA\_SCM} &\multicolumn{2}{c}{STA}&\multicolumn{2}{c}{DAOD} &\multicolumn{2}{c}{OURS} \\  
\cline{2-13}
&OS&OS*&OS&OS*&OS&OS*&OS&OS*&OS&OS*&OS&OS*\\
\hline
\textit{P1} $\rightarrow$ \textit{P2}  &41.9&44.0  &64.2&66.6  &60.7&60.9  &54.2 &55.0  &56.5&57.3 &\textbf{76.4}&\textbf{78.1} \\
\textit{P1} $\rightarrow$ \textit{P3}  &53.6&56.3  &66.4&69.1  &65.7&66.0  &67.7 &68.8  &52.2&53.1 &\textbf{75.7}&\textbf{77.4} \\
\textit{P1} $\rightarrow$ \textit{P4}  &64.6&67.9  &76.2&80.0  &79.5&80.3  &81.6 &83.6  &82.4&85.2 &\textbf{89.6}&\textbf{91.6} \\
\textit{P1} $\rightarrow$ \textit{P5}  &43.3&45.4  &49.1&50.2  &45.7&45.3  &42.4 &41.7  &46.1&47.3 &\textbf{57.2}&\textbf{58.0} \\
\textit{P2} $\rightarrow$ \textit{P1}  &56.7&59.5  &52.9&54.2  &63.6&65.2  &51.0 &51.6  &68.1&69.7 &\textbf{81.6}&\textbf{83.9} \\
\textit{P2} $\rightarrow$ \textit{P3}  &53.6&56.3  &61.5&63.5  &66.9&68.5  &58.3 &59.0  &69.9&71.7 &\textbf{76.5}&\textbf{78.3} \\
\textit{P2} $\rightarrow$ \textit{P4}  &73.5&77.1  &90.4&92.9  &91.2&93.6  &78.6 &80.6  &88.2&91.2 &\textbf{94.0}&\textbf{96.4} \\
\textit{P2} $\rightarrow$ \textit{P5}  &34.9&36.7  &45.1&45.9  &45.3&46.0  &39.6 &39.6  &49.4&49.8 &\textbf{51.8}&\textbf{52.6} \\
\textit{P3} $\rightarrow$ \textit{P1}  &66.9&68.4  &61.3&61.0  &75.2&77.3  &69.2 &70.7  &66.6&68.3 &\textbf{82.7}&\textbf{85.0} \\
\textit{P3} $\rightarrow$ \textit{P2}  &52.4&55.0  &64.1&64.6  &68.9&70.7  &59.5 &61.0  &68.5&70.4 &\textbf{76.0}&\textbf{78.0} \\
\textit{P3} $\rightarrow$ \textit{P4}  &70.5&74.0  &74.7&76.9  &\textbf{86.6}&\textbf{89.1}  &77.6 &79.8  &83.9&87.1 &{84.9}&{87.2} \\
\textit{P3} $\rightarrow$ \textit{P5}  &44.8&47.1  &46.3&46.7  &59.7&61.0  &46.3 &46.7  &52.3&53.3 &\textbf{62.8}&\textbf{64.2} \\
\textit{P4} $\rightarrow$ \textit{P1}  &63.7&66.8  &67.2&68.7  &85.7&86.9  &84.4 &86.6  &84.4&87.1 &\textbf{93.1}&\textbf{95.4} \\ 
\textit{P4} $\rightarrow$ \textit{P2}  &74.4&78.1  &82.2&85.0  &90.0&91.3  &89.7 &92.5  &82.4&84.8 &\textbf{93.9}&\textbf{96.2} \\
\textit{P4} $\rightarrow$ \textit{P3}  &58.7&61.7  &66.9&67.6  &\textbf{86.0}&\textbf{87.1}  &81.6 &84.4  &77.6&80.0 &{85.1}&{86.9} \\
\textit{P4} $\rightarrow$ \textit{P5}  &46.2&48.5  &61.7&63.8  &63.2&63.6  &68.8 &71.0  &59.9&61.3 &\textbf{71.3}&\textbf{72.7} \\
\textit{P5} $\rightarrow$ \textit{P1}  &30.2&23.5  &\textbf{64.2}&\textbf{66.6}  &54.3&55.7  &61.2 &62.6  &59.2&60.6 &{62.8}&{64.3} \\
\textit{P5} $\rightarrow$ \textit{P2}  &34.9&36.7  &35.4&35.8  &48.8&49.7  &49.8 &50.0  &35.0&34.8 &\textbf{50.2}&\textbf{51.1 }\\
\textit{P5} $\rightarrow$ \textit{P3}  &39.9&41.9  &45.1&46.3  &58.7&60.0  &46.5 &46.3  &44.6&44.4 &\textbf{69.2}&\textbf{70.8} \\
\textit{P5} $\rightarrow$ \textit{P4}  &55.8&58.6  &52.2&53.5  &71.1&73.0  &70.2 &71.7  &68.6&70.3 &\textbf{80.2}&\textbf{82.4} \\
\hline
Average            &53.0&55.2  &61.4&62.9  &68.3&69.6  &63.9 &65.2  &64.8&66.4 &\textbf{75.8}&\textbf{77.5}\\
\hline
\end{tabular}
\vspace{-1em}
\end{table*}

\begin{table*}[htbp]
\centering
\caption{\label{tab:ablation} Ablation study on \textit{Office-31}}
\begin{tabular}{p{1.25cm}p{0.7cm}<{\centering}p{0.7cm}<{\centering}p{0.7cm}<{\centering}p{0.7cm}<{\centering}p{0.7cm}<{\centering}p{0.7cm}<{\centering}p{0.7cm}<{\centering}p{0.7cm}<{\centering}p{0.7cm}<{\centering}p{0.7cm}<{\centering}p{0.7cm}<{\centering}p{0.7cm}<{\centering}p{0.7cm}<{\centering}p{0.7cm}}
\hline
% \multirow{2}{*}{Dataset}  &\multicolumn{2}{c}{AD} \\ 
\multirow{2}{*}{Dataset}  &\multicolumn{2}{c}{\textit{A} $\rightarrow$ \textit{D}}  &\multicolumn{2}{c}{\textit{A} $\rightarrow$ \textit{W}}  &\multicolumn{2}{c}{\textit{D} $\rightarrow$ \textit{A}} &\multicolumn{2}{c}{\textit{D} $\rightarrow$ \textit{W}}  &\multicolumn{2}{c}{\textit{W} $\rightarrow$ \textit{A}}  &\multicolumn{2}{c}{\textit{W} $\rightarrow$ \textit{D}} &\multicolumn{2}{c}{Avg} \\
\cline{2-15}
&OS&OS*&OS&OS*&OS&OS*&OS&OS*&OS&OS*&OS&OS*&OS&OS*\\
\hline
BADA                   &85.8&85.8 &76.9&76.6 &\textbf{89.4}&\textbf{91.5} &96.0&96.6  &{83.4}&83.1 &97.1&97.3  &88.0&88.5\\
BADA+$\Delta$          &92.7&93.3 &89.8&90.6 &81.6&81.7 &98.0&99.5  &\textbf{83.6}&78.9 &98.5&\textbf{100.0} &89.9&90.7\\
BADA+c                 &92.2&94.1 &87.6&89.0 &81.5&84.1 &97.7&\textbf{100.0} &80.3&83.4 &97.3&\textbf{100.0} &89.5&91.8\\
% OSBP+$\Delta$+c1       &94.0&94.6 &88.6&89.1 &83.2&83.5 &98.5&100.0 &83.1&81.6 &98.6&100.0 &90.7&91.5\\
% OSBP+$\Delta$+c2       &94.1&94.7 &89.5&90.2 &83.4&83.6 &98.3&99.8  &83.3&81.6 &98.6&100.0 &90.9&91.7\\
BADA+$\Delta$+c        &94.1&94.6 &89.2&89.7 &83.2&83.4 &\textbf{98.5}&\textbf{100.0} &83.3&81.9 &\textbf{98.6}&\textbf{100.0} &90.9&91.7\\
BADA+$\Delta_\epsilon $    &95.5&97.0 &\textbf{92.6}&\textbf{94.0} &82.3&82.6 &98.0&99.5  &{83.4}&79.5 &98.4&\textbf{100.0} &91.0&92.2\\
% OSBP+$\Delta_\epsilon $+c1 &96.6&98.1 &92.5&93.8 &83.9&84.6 &98.3&99.8  &82.6&81.9 &98.6&100.0 &92.0&93.1\\
% OSBP+$\Delta_\epsilon $+c2 &96.3&97.9 &92.5&94.0 &84.1&84.8 &98.3&99.8  &82.6&82.4 &98.5&100.0 &92.0&93.3\\
OURS                   &\textbf{96.0}&\textbf{97.5} &92.5&93.7 &{85.3}&{86.0} &98.4&\textbf{100.0} &{83.2}&\textbf{83.9} &\textbf{98.6}&\textbf{100.0} &\textbf{92.3}&\textbf{93.5}\\
\hline
\end{tabular}
\vspace{-1em}
\end{table*}

Results on three tasks of \textit{Digit} datasets are shown in Table \ref{tab:digit}, Obviously, our method achieves the best performance ($89.0\%$ on {OS} and $90.0\%$ on {OS*}) within three tasks. Moreover, compared to \textit{U} $\rightarrow$ \textit{M} and \textit{M} $\rightarrow$ \textit{U}, \textit{M} $\rightarrow$ \textit{U} is more challenging. There is a bigger distribution  between \textit{S} and \textit{M}. Whereas on the most difficult task, our method still outperforms the best baseline STA by $6\%$ and $7.2\%$ on {OS} and {OS*} respectively. It is worth noting that DAOD is a shallow method, which cannot extract feature by convolutional neural network. Therefore there is no comparison on Digits. The results of ATI-$\lambda$ are from \cite{liu2019separate}.

Results on standard benchmark object datasets (\textit{Office-31} and \textit{Office-Home}) are recorded in Table \ref{tab:office}. For \textit{Office-31}, our method significantly outperforms baselines among $4$ out of $6$ transfer tasks. Especially on \textit{A} $\rightarrow$ \textit{D}, our method surpasses the most competitive baseline SCA\_SCM by $5.9\%$ and $5.5\%$ on {OS} and {OS*} respectively. For \textit{Office-Home}, our method also achieves better performance than baselines among $9$ out of $12$ transfer tasks.

Results on \textit{PIE} datasets are shown in Table~\ref{tab:pie}. Although \textit{PIE} is a dataset with significant distributional discrepancy, our method still outperforms baselines among $17$ out of $20$ transfer tasks. Specifically, our method surpasses the best baseline SCA\_SCM by $7.5\%$ and $7.9\%$ on {OS}, {OS*} respectively.

Moreover, we observe that: 1) the performance of ATI-$\lambda$ is lower than that of other methods. That is because ATI-$\lambda$ cannot accurately separate unknown data, and it needs numerous unknown data in the source domain to train a classifier to recognize unknown data.
2) OSBP and SCA\_SCM leverage an adversarial net to separate unknown data, which can find a rough boundary between known classes and unknown classes. However, the classifier is easily affected by hyper-parameter t, which means that the classifier cannot recognize the target data well. For example, for OSBP, in \textit{Digits}, the accuracy of classifying unknown data is significantly higher than known classes, but the opposite situation is apparent in \textit{Office-31}, which proves that this method is not robust. For SCA\_SCM, it cannot recognize unknown data well. Especially on the task \textit{D} $\rightarrow$ \textit{A} of \textit{Office-31}, SCA\_SCM fails to recognize unknown data. That is because {OS*} is greater than {OS} by $8.6\%$. 4) STA separates known data and unknown data by a multi binary classifier. It can achieve a good performance in known classes, but it cannot cope with unknown classes well, especially for the tasks with a large domain gap. 
% 5) Due to the defect of open set difference, DAOD easily fails to align distribution, which results in a dissatisfied performance.

Compared to baselines, the proposed risk estimator, $\epsilon$-open set difference, can help us effectively estimate the risk of the classifier on unknown data. As a result, a clear boundary between known classes and unknown classes can be found. Moreover, our method leverages a novel open-set conditional adversarial strategy to capture the multimodal structure of distributions, which can be used to align distribution adequately. Better recognizing unknown data and better aligning distribution make our method achieve an excellent performance, which is the reason why we can outperform all baselines on $4$ benchmark datasets.

\begin{figure*}[!t]
    \centering
    \includegraphics[scale=0.38, trim=10 10 0 0, clip]{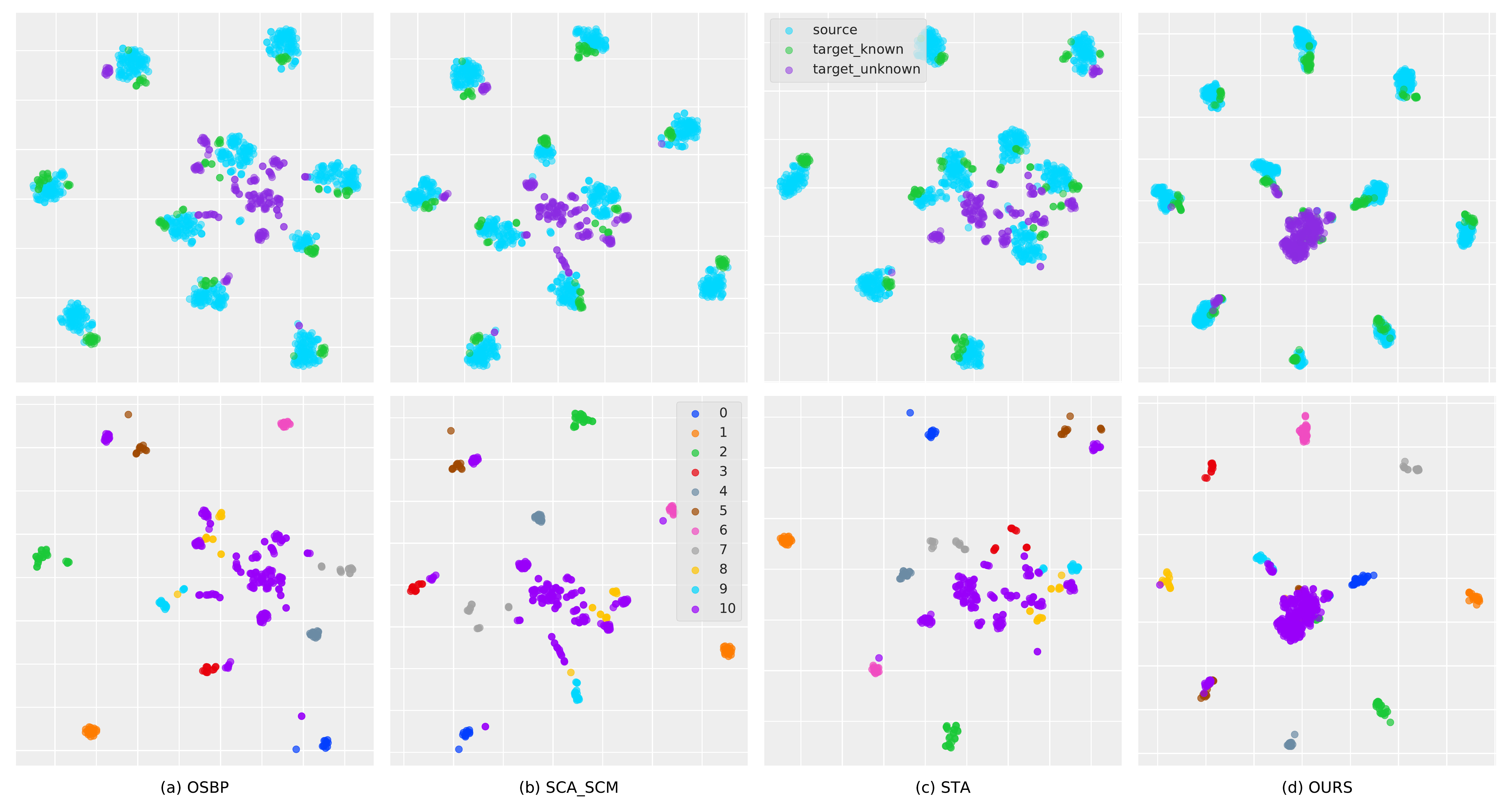}
    \caption{Feature visualization on \textit{A} $\rightarrow$ \textit{D}. \textbf{First row}: visualization of target and source features. \textit{Blue points} indicate source samples. \textit{Green points} indicate target known samples. \textit{Purple points} indicate target unknown samples. \textbf{Second row}: visualization of target samples only.}
    \label{fig:tsne1}
\end{figure*}

\subsection{Analysis}
\subsubsection{Ablation Study}
It is necessary to conduct the ablation experiments to demonstrate the effect of each part of our method. Since our method is based on binary adversarial domain adaptation (BADA) \cite{saito2018open}, we introduce open set difference ($\Delta$), $\epsilon$-open set difference ($\Delta\epsilon $), and conditional adversarial domain adaptation (c) into {BADA} and construct ablation experiments as follows: (1) {BADA}, (2) {BADA}+$\Delta$, (3) {BADA}+c, (4) {BADA}+$\Delta$+c, (5) {BADA}+$\Delta_\epsilon$ and (6) {OURS} (\textit{i.e.}, {BADA}+$\Delta_\epsilon $+c). Results of ablation experiments are shown in table \ref{tab:ablation}.

From Table \ref{tab:ablation}, the following facts can be verified: 1) by comparing {BADA}, {BADA}+c, and {BADA}+$\Delta_\epsilon$, the accuracy of {BADA} is the lowest, which proves that $\Delta_\epsilon$ and conditional adversarial domain adaptation are all useful for UOSDA; 2) the results of {BADA}+$\Delta_\epsilon $ and {OURS} is higher than {BADA}+$\Delta$ and {BADA}+$\Delta$+c respectively, which adequately indicates that $\Delta_\epsilon $ can overcome the issue caused by $\Delta$. The method with $\Delta_\epsilon $ can establish a boundary between known and unknown classes, preventing the negative transfer caused by unknown classes during the process of distribution alignment; 3) the accuracies of {BADA}+$\Delta$+c and OURS are higher than those of {BADA}+$\Delta$ and {BADA}+$\Delta_\epsilon$ respectively, which proves that the novel conditional adversarial domain adaptation effectively elevates the performance of our method.

\subsubsection{Visualization}
In order to intuitively demonstrate the effect of our method, we visualize the 2D features of source and target by $t$-SNE \cite{maaten2008visualizing}, which is an effective dimensionality reduction method. Fig. \ref{fig:tsne1} shows the effect of domain adaptation of baselines and our method. Clearly, our method outperforms baselines in separating unknown data and aligning distributions of two domains.

From the first row of Fig. \ref{fig:tsne1}, OSBP, SCA\_SCM and STA cannot adequately align distributions of source and target domains, which is because they cannot distinguish unknown data from known data. As a result, the distribution of unknown classes gets closer to the distribution of known classes. However, our method, in Fig. \ref{fig:tsne1}(d), can effectively recognize unknown data and make the distribution of unknown classes far from the distribution of known classes. The second row shows the feature distribution of the target data only. Compared to baselines, it is clear that our method can effectively recognize unknown data and align distributions.

\begin{figure*}[ht]
    \centering
    \includegraphics[scale=0.44, trim=0 10 0 0, clip]{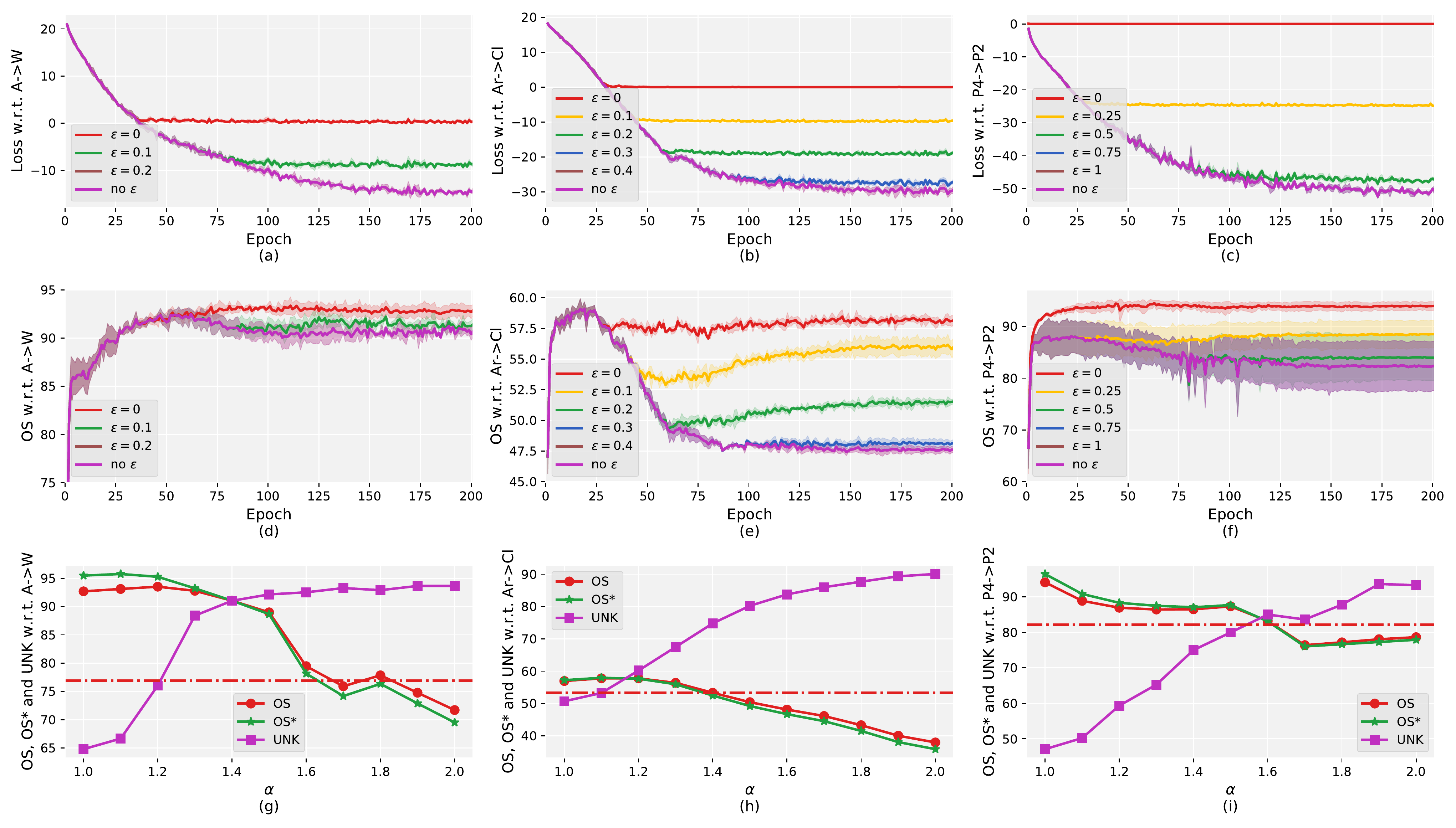}
    \caption[Caption for LOF]{Parameter analyses \textit{w.r.t.} $\epsilon $ and $\alpha$. Experiments are conducted on \textit{A} $\rightarrow$ \textit{W} of \textit{Office-31} (first column), \textit{Ar} $\rightarrow$ \textit{Pr} of \textit{Office-Home} (second column) and \textit{P4} $\rightarrow$ \textit{P2} (third column). First row: the value of $\Delta_\epsilon $ or $\Delta$ (``no $\epsilon$" indicates $\Delta$). Second row: The accuracy of {OS} \textit{w.r.t.} $\Delta_\epsilon$ and $\Delta$ when $\epsilon $ changes. Third row: The accuracy of {OS}, OS*, and {UNK} \textit{w.r.t.} $\alpha$. The losses in (a), (b) and (c) are the values of $\Delta$ or $\Delta_\epsilon$. It is worth noting that: The line of ``$\epsilon$=0.2" coincides with the line of ``no $\epsilon$" in (a) and (d); The line of ``$\epsilon$=0.4" coincides with the line of ``no $\epsilon$" in (b) and (e); The lines of ``$\epsilon$=0.75" and ``$\epsilon$=1" coincide with the line of ``no $\epsilon$" in (c) and (f).}
    \label{fig:beta_ana}
\end{figure*}

% \footnote{\textit{UNK} is a metric to evaluate the accuracy on unknown target samples, which is defined as.} {\rm Acc(UNK) }= \frac{|\mathbf{x}\in\mathbf{X}_{K+1}^t\bigwedge {\bm C}\circ{\bm G}(\mathbf{x})=K+1|}{|\mathbf{X}^t_{K+1}|}

\subsubsection{Analysis on $\epsilon$-open Set Difference}
In this paper, there are two critical parameters in $\epsilon$-open set difference: $\epsilon $ and $\alpha$.
Theoretically, $\epsilon $ is a variable related to distributional discrepancy. According to Eq. \eqref{lowerbound}, distributional discrepancy is greater than the negative open set difference. We hope that distributional discrepancy is close to $0$. Thus an intuitive thought is to set $\epsilon $ as $0$. Moreover, $\alpha$ is equal to $1-\pi_{K+1}^t$, but $\pi_{K+1}^t$ is an unknown value and is hard to estimate in a batch for a deep method. Therefore, we conduct related experiments to demonstrate the effect of these two parameters for our method. Fig. \ref{fig:beta_ana} shows the effect of parameters on \textit{A} $\rightarrow$ \textit{W} of \textit{Office-31} (first column), \textit{Ar} $\rightarrow$ \textit{Cl} of \textit{Office-Home} (second column) and \textit{P4} $\rightarrow$ \textit{P2} (third column).

The influence of $\epsilon $ is shown in the first and second rows. On the task \textit{A} $\rightarrow$ \textit{W}, the accuracy of {OS} is decreasing with the increase of $\epsilon$, which is because $\epsilon $ is related to distribution alignment. The bigger $\epsilon $ means the smaller value of the lower bound of distributional discrepancy. However, it is worth noting that the value of {OS} does not change, which is because the domain gap between \textit{A} and \textit{W} is small. Therefore the effect of $\Delta_\epsilon $ is same as $\Delta$ when $\epsilon $ greater than a constant. In Figs. \ref{fig:beta_ana}(a) and \ref{fig:beta_ana}(d), the line of $\epsilon $ equal to 0.2 coincides with the line of ``no $\epsilon$" (\textit{i.e. $\Delta$}).

On task \textit{Ar} $\rightarrow$ \textit{Cl}, there is a large domain gap. In the same way as task \textit{A} $\rightarrow$ \textit{W}, The bigger $\epsilon $, the smaller {OS}. The line of ``$\epsilon$=0.4" coincides with the line of $\Delta$, which also indicates that the distributional discrepancy of \textit{Office-Home} is larger than \textit{Office-31}. On task \textit{P4} $\rightarrow$ \textit{P2}, the domain gap is also large. These line ``$\epsilon$=$0.75$" and ``$\epsilon$=$1.0$" coincide with the line of $\Delta$. It is worth noting that the increasing tendency of the yellow line and the green line after the turning point in Fig. \ref{fig:beta_ana}(e), owes to the effect of $\Delta_\epsilon$ and it prevents the problem caused by open set difference.

The effect of parameter $\alpha$ is shown in the third row. The dash line denotes the accuracy of the {OSBP}. From Figs. \ref{fig:beta_ana}(c), \ref{fig:beta_ana}(f) and \ref{fig:beta_ana}(i), we can conclude that: When we choose a large $\alpha$, the classifier tends to recognize data as unknown, which leads to the increase of accuracy on unknown classes ({UNK}\footnote{{UNK} is a metric to evaluate the accuracy on unknown target data \cite{saito2018open}.}) and the decrease of accuracy on known classes (OS*). When we choose a small $\alpha$, the classifier tends to distinguish data as known data. That is why the classifier achieves a good performance on {OS*} and a bad performance on UNK.
On Fig. \ref{fig:beta_ana}(g), it is easy to observe that our method can outperform the best baselines when $\alpha \in [1.0, 1.6]$. So we recommend to set $\alpha$ in range of $[1.0, 1.6]$ on \textit{Office-31}. Similarly, the recommendation parameter range is $[1.0, 1.4]$ and $[1.0, 1.6]$ on \textit{Office-Home} and \textit{PIE} respectively. Thus we recommend to set  $\alpha$ in range of $[1.0, 1.4]$.

\section{Conclusion and Future Work}
In this paper, we tackled a challenging problem called unsupervised open set domain adaptation (UOSDA). We proposed a practical theoretical bound for UOSDA, which contains an effective risk estimator ($\Delta_\epsilon $) to evaluate the risk on data with unknown classes. Furthermore, we proposed a DNN-based UOSDA method under the guidance of the proposed theoretical bound. The method can accurately estimate the risk of the classifier on data with unknown classes via $\Delta_\epsilon$ and adequately align the distributions of data with known classes via a novel open-set conditional adversarial training strategy. Experiments on several benchmark datasets demonstrated that our method significantly outperforms state-of-the-art UOSDA methods.

In the future, we aim to investigate a more challenging problem called universal domain adaptation \cite{longmingsheng1}, which contains unknown classes in both source and target domains. This setting is a more general one and includes UCSDA, UOSDA, and partial domain adaptation \cite{DBLP:conf/cvpr/CaoL0J18} as its special cases.

% \footnotetext{\textit{UNK} is a metric to evaluate the accuracy on unknown target samples, which is defined as ${\rm Acc(UNK) }= \frac{|\mathbf{x}\in\mathbf{X}_{K+1}^t\bigwedge {\bm C}\circ{\bm G}(\mathbf{x})=K+1|}{|\mathbf{X}^t_{K+1}|}$.}

% \appendices

% % you can choose not to have a title for an appendix
% % if you want by leaving the argument blank
% \section{}

\section*{Acknowledgment}
The work presented in this paper was supported by the Australian Research Council (ARC) under DP170101632 and FL190100149. The first author particularly thanks the support of UTS-CAI during his visit.

\ifCLASSOPTIONcaptionsoff
  \newpage
\fi

\bibliographystyle{IEEEtran}
\bibliography{references.bib}
\vspace{-1cm}
\begin{IEEEbiography}[{\includegraphics[width=1.0in,height=1.8in,clip,keepaspectratio]{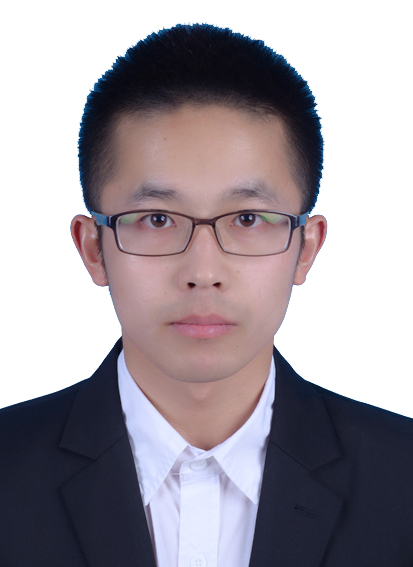}}]{Li Zhong} is currently working toward his M.Sc degree in control engineering with the Faculty of Information Science and Technology, Tsinghua University. He received his B.Sc degree in automation from the School of Electronic Control and Engineering, Changan University, China, in 2018. He is a member of the Intelligent Computing Lab, Tsinghua University. His research interests includes transfer learning and domain adaptation.
\end{IEEEbiography}
\vspace{-1.4cm}

\begin{IEEEbiography}[{\includegraphics[width=1.0in,height=1.8in,clip,keepaspectratio]{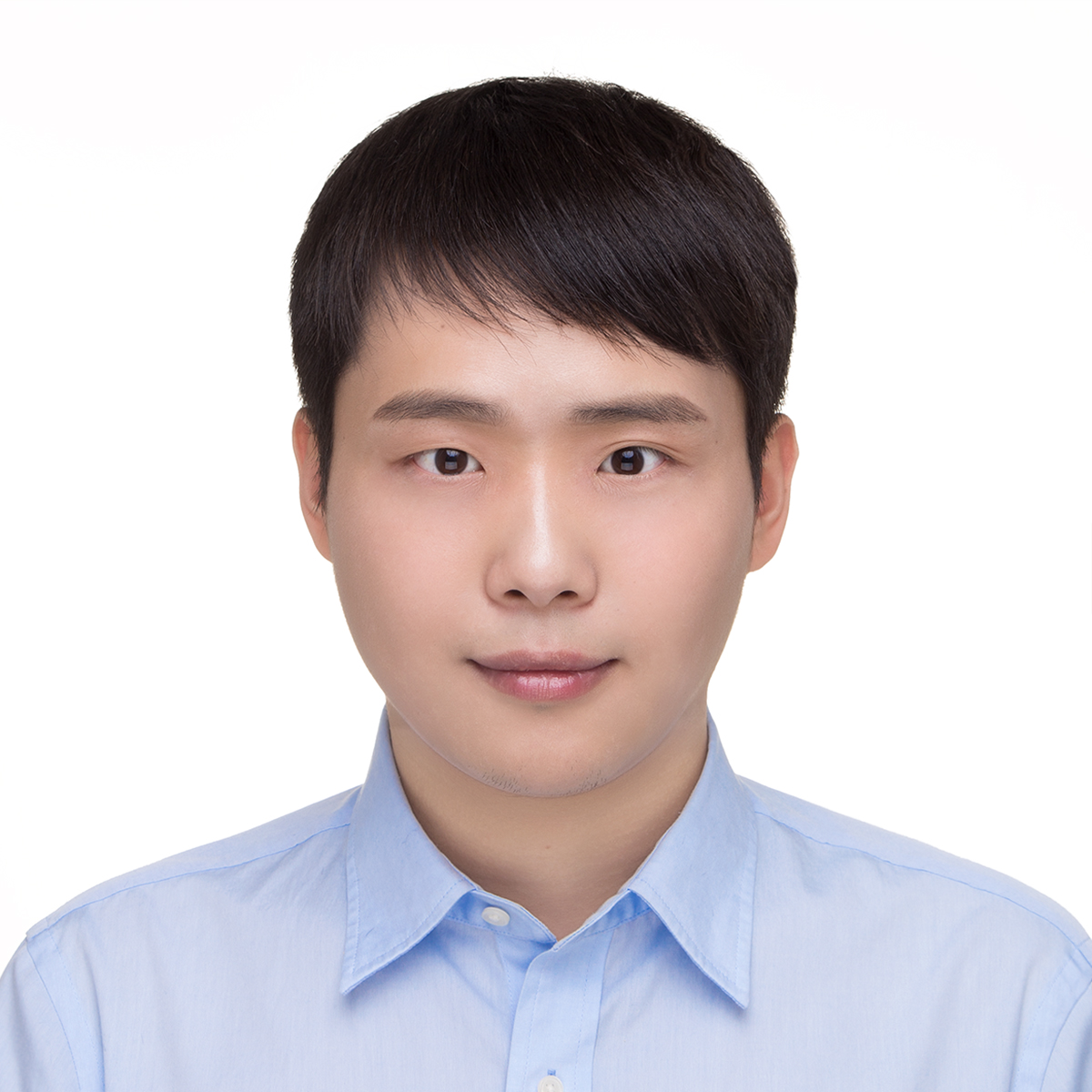}}]{Zhen Fang}
received his M.Sc. degree in pure mathematics from the School of Mathematical Sciences Xiamen University, Xiamen, China, in 2017. He is working toward a PhD degree with the Faculty of Engineering and Information Technology, University of Technology Sydney, Australia. His research interests include transfer learning and domain adaptation. He is a Member of the Decision Systems and e-Service Intelligence (DeSI) Research Laboratory, CAI, University of Technology Sydney.
\end{IEEEbiography}
\vspace{-1.4cm}

\begin{IEEEbiography}[{\includegraphics[width=1in,height=1.25in,clip,keepaspectratio]{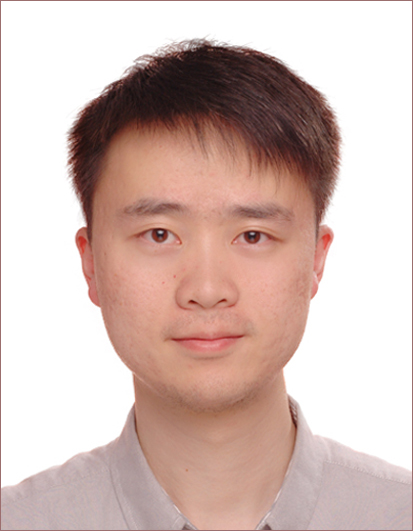}}]{Feng Liu}
is a Doctoral candidate in Centre for Artificial intelligence, Faculty of Engineering and Information Technology, University of Technology Sydney, Australia. He received an M.Sc. degree in probability and statistics and a B.Sc. degree in pure mathematics from the School of Mathematics and Statistics, Lanzhou University, China, in 2015 and 2013, respectively. His research interests include domain adaptation and two-sample test. He has served as a senior program committee member for ECAI and program committee members for NeurIPS, ICML, IJCAI, CIKM, FUZZ-IEEE, IJCNN and ISKE. He also served as reviewers for TPAMI, TNNLS, TFS and TCYB. He has received the UTS-FEIT HDR Research Excellence Award (2019), Best Student Paper Award of FUZZ-IEEE (2019) and UTS Research Publication Award (2018).
\end{IEEEbiography}
\vspace{-1.4cm}

\begin{IEEEbiography}[{\includegraphics[width=1in,height=1.25in,clip,keepaspectratio]{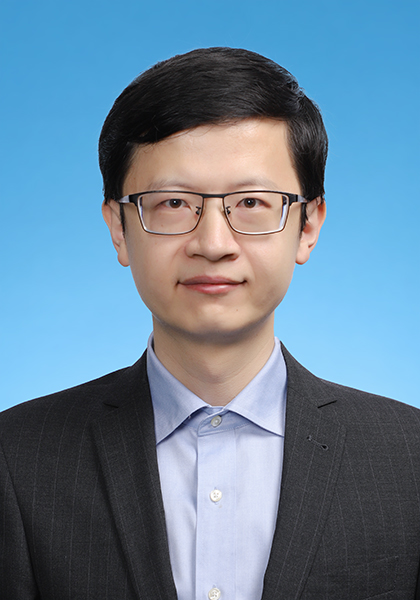}}]{Bo Yuan} received the B.E. degree from Nanjing University of Science and Technology, P.R.China, in 1998, and the M.Sc. and Ph.D. degrees from The University of Queensland (UQ), Australia, in 2002 and 2006, respectively, all in Computer Science. From 2006 to 2007, he was a Research Officer on a project funded by the Australian Research Council at UQ. He is currently an Associate Professor in the Division of Informatics, Shenzhen International Graduate School, Tsinghua University, and a member of the Intelligent Computing Lab. He is the author of more than 100 refereed research papers in data mining, evolutionary computation and GPU computing.
\end{IEEEbiography}
\vspace{-1.4cm}

\begin{IEEEbiography}[{\includegraphics[width=1in,height=1.25in,clip,keepaspectratio]{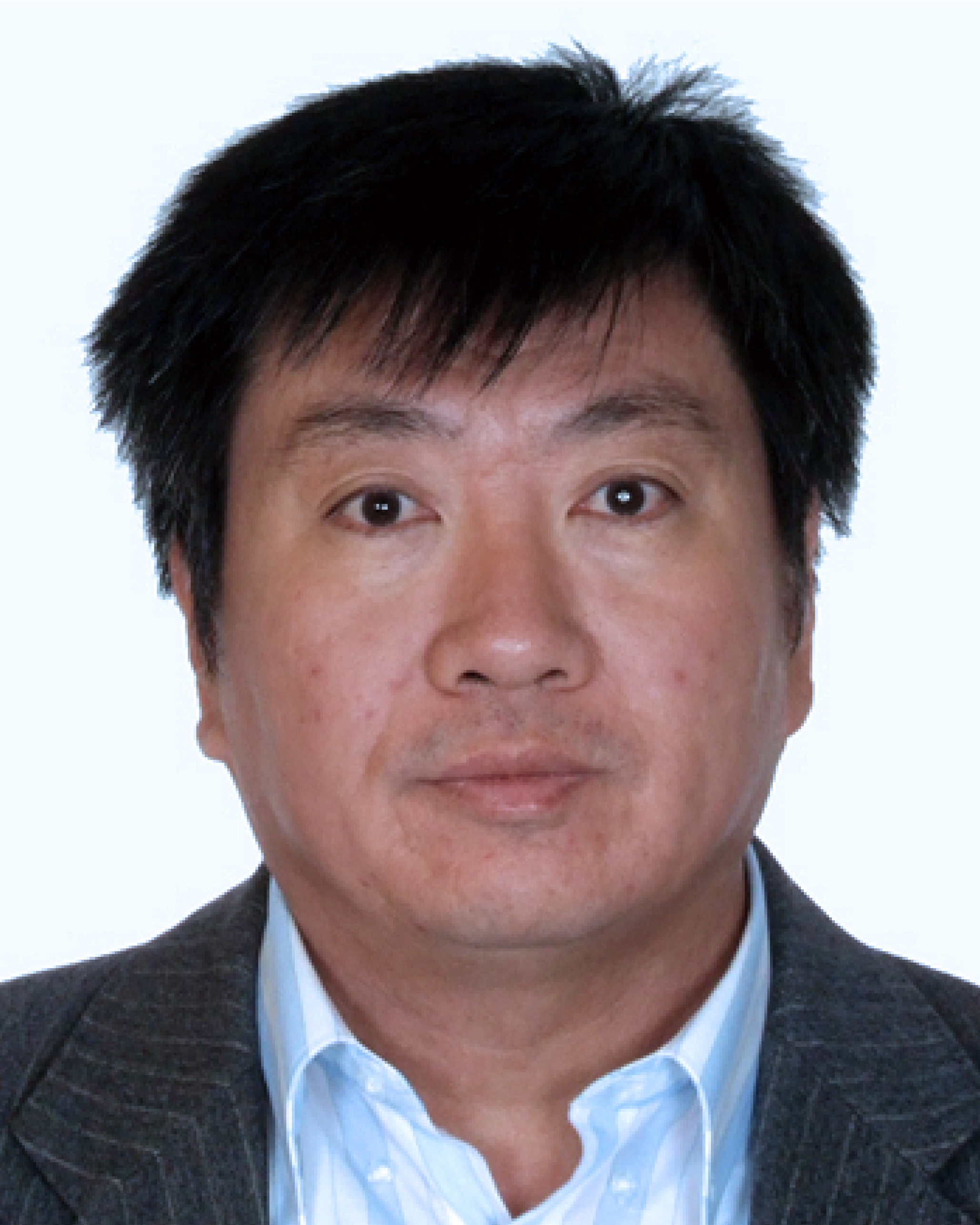}}]{Guangquan Zhang}
is a Professor and Director of the Decision Systems and e-Service Intelligent (DeSI) Research Laboratory, Faculty of Engineering and Information Technology, University of Technology Sydney, Australia. He received his PhD in applied mathematics from Curtin University of Technology, Australia, in 2001.
His research interests include fuzzy machine learning, fuzzy optimization, and machine learning and data analytics. He has authored four monographs, five textbooks, and 350 papers including 160 refereed international journal papers. Dr. Zhang has won seven Australian Research Council (ARC) Discovery Project grants and many other research grants. He was awarded an ARC QEII Fellowship in 2005. He has served as a member of the editorial boards of several international journals, as a guest editor of eight special issues for IEEE Transactions and other international journals, and has co-chaired several international conferences and work-shops in the area of fuzzy decision-making and knowledge engineering. 
\end{IEEEbiography}
\vspace{-10cm}

\begin{IEEEbiography}[{\includegraphics[width=1in,height=1.25in,clip,keepaspectratio]{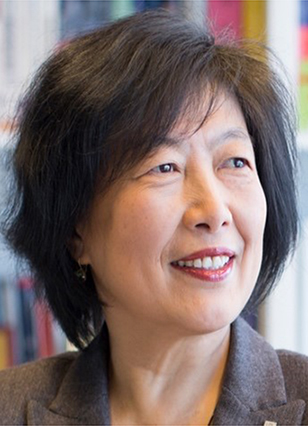}}]{Jie Lu} (F’18) is a Distinguished Professor and the Director of the Centre for Artificial Intelligence at the University of Technology Sydney, Australia. She received her PhD degree from Curtin University of Technology, Australia, in 2000. Her main research interests arein the areas of fuzzy transfer learning, concept drift, decision support systems, and recommender systems. She is an IEEE fellow, IFSA fellow and Australian Laureate fellow. She has published six research books and over 450 papers in refereed journals and conference proceedings; has won over 20 ARC Laureate, ARC Discovery Projects, government and industry projects. She serves as Editor-In-Chief for Knowledge-Based Systems (Elsevier) and Editor-In-Chief for International journal of computational intelligence systems. She has delivered over 25 keynote speeches at international conferences and chaired 15 international conferences. She has received various awards such as the UTS Medal for Research and Teaching Integration (2010), the UTS Medal for Research Excellence (2019), the Computer Journal Wilkes Award (2018), the IEEE Transactions on Fuzzy Systems Outstanding Paper Award (2019), and the Australian Most Innovative Engineer Award (2019).
\end{IEEEbiography}
\end{document}

\title{Bridging the Theoretical Bound and Deep Algorithms for Open Set Domain Adaptation}

\author{}
\maketitle


\tableofcontents
\newpage
\section*{\large{SUMMARY}}
The proof for Theorem 1 is provided in Appendix A. The details of network architecture is demonstrated in Appendix B.

\section{\large{Appendix A: Proof of Theorem 1 }}

\begin{theorem}\label{-1000}
Given a feature transformation ${\hm G}: \mathcal{X}\rightarrow \mathcal{X}_{\hm G}$, a  loss function $\ell$ satisfying conditions 1-3 introduced in Section III-B-2), a nonegative constant $\epsilon $ and  a hypothesis $\mathcal{H}_{\hm G}\subset\{{\hm C}:\mathcal{X}_{\hm G}\rightarrow \mathcal{Y}^t\}$ with a mild condition that the constant vector value function ${{\hm g}}:={\mathbf{y}}_{C+1}\in \mathcal{H}_{\hm G}$, then for any ${\hm C}\in \mathcal{H}_{\hm G}$,  we have 
\begin{equation}
\label{eq: new_ub}
\begin{split}
 ~~&\frac{L^t({\hm C}\circ {\hm G})}{1-\pi_{K+1}^t}\leq   \overbrace{L^s({\hm C}\circ {\hm G})}^{\text{{Source Risk}}} +{\overbrace{2d_{\Delta_{{\hm C},{\hm G}}}^{\ell}({\otimes_{{\hm C}}}_{\#}P_{X^s},{\otimes_{{\hm C}}}_{\#}P_{X^t|\mathcal{Y}^s})}^{\text{{Tensor distributional discrepancy}}}}\\&+\underbrace{\max\{-\epsilon , \frac{L^t_{u,K+1}({\hm C}\circ {\hm G})}{1-\pi_{K+1}^t}
 -L^s_{u,K+1}({\hm C}\circ {\hm G})\}}_{\text{{Open Set Difference}}~\Delta_\epsilon }+\Lambda,
 \end{split}
\end{equation}
 where $L^s({\hm C}\circ {\hm G})$ and $L^t({\hm C}\circ {\hm G})$ are the risks defined in (5) in Section III, $L^s_{u,K+1}({\hm C}\circ {\hm G})$ and $L^t_{u,K+1}({\hm C}\circ {\hm G})$ are the risks defined in (8) in Section III, $L^t_{*}({\hm C}\circ {\hm G})$ is the partial risk defined in (6) in Section III and $\Lambda=\underset{{\hm C}\in \mathcal{H}_{\hm G}}{\min}  ~L^s({\hm C}\circ {\hm G})+L^t_{*}({\hm C}\circ {\hm G})$. 
\end{theorem}

\begin{proof}

Firstly, we introduce the basic idea.
 It is easy to check that
\begin{equation*}
\begin{split}
    L^t({\hm C}\circ {\hm G})=(1-\pi_{K+1}^t)L^t_*({\bm h})+\pi_{K+1}^tL^t_{K+1}({\hm C}\circ {\hm G}),
\end{split}
\end{equation*}
then
\begin{equation}\label{Th1}
\begin{split}
    \frac{L^t({\hm C}\circ {\hm G})}{1-\pi_{K+1}^t}-L^s({\hm C}\circ {\hm G})=L^t_*({\hm C}\circ {\hm G})-L^s({\hm C}\circ {\hm G})+\frac{\pi_{K+1}^t}{1-\pi_{K+1}^t}L^t_{K+1}({\hm C}\circ {\hm G}),
\end{split}
\end{equation}
thus we separate the proof into two main steps. For the first step, we mainly consider that  $L^t_{*}({\hm C}\circ {\hm G})-L^s({\hm C}\circ {\hm G})$. For the second step, we investigate $L^t_{K+1}({\hm C}\circ {\hm G})$.
\\
If we can prove that
\begin{equation}\label{Th2}
   L^t_{*}({\hm C}\circ {\hm G})-L^s({\hm C}\circ {\hm G})\leq\Lambda +d_{\Delta_{{\hm C},{\hm G}}}^{\ell}({\otimes_{{\hm C}}}_{\#}P_{X^s},{\otimes_{{\hm C}}}_{\#}P_{X^t|\mathcal{Y}^s}), 
\end{equation}
and
 \begin{equation}\label{Th3}
 \begin{split}
     \frac{\pi_{K+1}^t}{1-\pi_{K+1}^t}L^t_{K+1}({\hm C}\circ {\hm G})\leq & {d_{\Delta_{{\hm C},{\hm G}}}^{\ell}({\otimes_{{\hm C}}}_{\#}P_{X^s},{\otimes_{{\hm C}}}_{\#}P_{X^t|\mathcal{Y}^s})}+\max \{-\epsilon,\frac{L^t_{u,K+1}({\hm C}\circ {\hm G})}{1-\pi_{K+1}^t}-L^t_{u,K+1}({\hm C}\circ {\hm G})\},
     \end{split}
 \end{equation}
 then combining  (\ref{Th2}),  (\ref{Th3}) with (\ref{Th1}), we have
 \begin{equation*}
    \frac{L^t({\hm C}\circ {\hm G})}{1-\pi_{K+1}^t}\leq   {L^s({\hm C}\circ {\hm G})} +{2{d_{\Delta_{{\hm C},{\hm G}}}^{\ell}({\otimes_{{\hm C}}}_{\#}P_{X^s},{\otimes_{{\hm C}}}_{\#}P_{X^t|\mathcal{Y}^s})}}+\Lambda+\Delta_\epsilon.
\end{equation*}
%\begin{equation}
%\begin{split}
%d^{\ell}_{\Delta_{{\hm C},{\hm G}}}({\otimes_{{\hm C}}}_{\#}P_{X^s},{\otimes_{{\hm C}}}_{\#}P_{X^t|\mathcal{Y}^s})&=2\sup_{\delta\in \Delta_{{\hm C},{\hm G}}}\Big| \underset{{{\mathbf{z}}\sim \otimes_{{\hm C}}}_{\#}P_{X^s}}{\mathbb{E}}1_{\delta(\mathbf{z})}-\underset{{{\mathbf{z}}\sim \otimes_{{\hm C}}}_{\#}P_{X^t|\mathcal{Y}^s}}{\mathbb{E}}1_{\delta(\mathbf{z})}\Big|
%\end{split}
%\end{equation}

\textbf{Step 1.} we claim that $ L^t_{*}({\hm C}\circ {\hm G})-L^s({\hm C}\circ {\hm G})\leq\Lambda +d_{\Delta_{{\hm C},{\hm G}}}^{\ell}({\otimes_{{\hm C}}}_{\#}P_{X^s},{\otimes_{{\hm C}}}_{\#}P_{X^t|\mathcal{Y}^s})$.

\begin{equation}\label{l1}
	\begin{split}
		~~~L^t_{*}({\hm C}\circ {\hm G})-L^s({\hm C}\circ {\hm G})=&\int_{\mathcal{X}\times \mathcal{Y}^t} \ell({\hm C}\circ {\hm G}({\mathbf{x}}),{\mathbf{y}}){\rm d} P_{X^tY^t|\mathcal{Y}^s}-\int_{\mathcal{X}\times \mathcal{Y}^s} \ell({\hm C}\circ {\hm G}({\mathbf{x}}),{\mathbf{y}}){\rm d} P_{X^sY^s}\\{\leq}&L^t_*(\tilde{\hm C}\circ {\hm G})+\int_{\mathcal{X}\times \mathcal{Y}^t} \ell({\hm C}\circ {\hm G}({\mathbf{x}}),\tilde{\hm C}\circ {\hm G}({\mathbf{x}})){\rm d} P_{X^tY^t|\mathcal{Y}^s}\\+&L^s(\tilde{\hm C}\circ {\hm G})-\int_{\mathcal{X}\times \mathcal{Y}^s} \ell({\hm C}\circ {\hm G}({\mathbf{x}}),\tilde{\hm C}\circ {\hm G}({\mathbf{x}})){\rm d} P_{X^sY^s},
	\end{split}
\end{equation}
where $\tilde{\hm C}$ is any function from $\mathcal{H}_{\hm G}$.

Then we note that
\begin{equation}
\begin{split}
	&~~~d^{\ell}_{\Delta_{{\hm C},{\hm G}}}({\otimes_{{\hm C}}}_{\#}P_{X^s},{\otimes_{{\hm C}}}_{\#}P_{X^t|\mathcal{Y}^s})=\sup_{\tilde{C}\in \mathcal{H}_{\hm G}} \left |\int_{\mathcal{X}} \ell({\hm C}\circ {\hm G} (\mathbf{x}), \tilde{\hm C}\circ {\hm G}(\mathbf{x})){\rm d}P_{X^s} -\int_{\mathcal{X}} \ell({\hm C}\circ {\hm G} (\mathbf{x}), \tilde{\hm C}\circ {\hm G}(\mathbf{x})){\rm d}P_{X^t|\mathcal{Y}^s} \right |.
\end{split}
\end{equation}
This is because, according to conditions 1-3, we have that
\begin{equation}
	\begin{split}
	\underset{{{\mathbf{z}}\sim \otimes_{{\hm C}}}_{\#}P_{X^s}}{\mathbb{E}}{{\rm sgn}\circ \delta_{\widetilde{\hm C}}(\mathbf{z})}&=\int {{\rm sgn}\circ\delta_{\widetilde{\hm C}}(\mathbf{z})} {\rm d} {\otimes_{{\hm C}}}_{\#}P_{X^s}\\&=\int_{\mathcal{X}}  {|{\hm C}\circ {\hm G} (\mathbf{x}) - \widetilde{\hm C}\circ {\hm G}(\mathbf{x})|}  {\rm d}P_{X^s}\\&=\int_{\mathcal{X}}  \ell({\hm C}\circ {\hm G} (\mathbf{x}), \widetilde{\hm C}\circ {\hm G}(\mathbf{x})){\rm d}P_{X^s},
	\end{split}
\end{equation}
similarly, 
\begin{equation}
\begin{split}
\underset{{{\mathbf{z}}\sim \otimes_{{\hm C}}}_{\#}P_{X^t|\mathcal{Y}^s}}{\mathbb{E}}{{\rm sgn}\circ\delta_{\widetilde{\hm C}}(\mathbf{z})}=\int_{\mathcal{X}}  \ell({\hm C}\circ {\hm G} (\mathbf{x}), \widetilde{\hm C}\circ {\hm G}(\mathbf{x})){\rm d}P_{X^t|\mathcal{Y}^s}
\end{split}
\end{equation}
hence, according to the definition of $d^{\ell}_{\Delta_{{\hm C},{\hm G}}}({\otimes_{{\hm C}}}_{\#}P_{X^s},{\otimes_{{\hm C}}}_{\#}P_{X^t|\mathcal{Y}^s})$, we have
\begin{equation}\label{l2}
\begin{split}
	&~~~d^{\ell}_{\Delta_{{\hm C},{\hm G}}}({\otimes_{{\hm C}}}_{\#}P_{X^s},{\otimes_{{\hm C}}}_{\#}P_{X^t|\mathcal{Y}^s})=\sup_{\tilde{C}, {G} \in \Delta_{{\hm C},{\hm G}}} \left |\int_{\mathcal{X}} \ell({\hm C}\circ {\hm G} (\mathbf{x}), \tilde{\hm C}\circ {\hm G}(\mathbf{x})){\rm d}P_{X^s} -\int_{\mathcal{X}} \ell({\hm C}\circ {\hm G} (\mathbf{x}), \tilde{\hm C}\circ {\hm G}(\mathbf{x})){\rm d}P_{X^t|\mathcal{Y}^s} \right |.
\end{split}
\end{equation}
Combining (\ref{l1}) and (\ref{l2}), we have
\begin{equation}\label{l4}
\begin{split}
L^t_{*}({\hm C}\circ {\hm G})-L^s({\hm C}\circ {\hm G})\leq &\underset{\tilde{C}\in \Delta_{C,G}}{\min} ~\left(L^t_{*}(\tilde{\hm C}\circ {\hm G})+L^s(\tilde{\hm C}\circ {\hm G})\right)+d^{\ell}_{\Delta_{{\hm C},{\hm G}}}({\otimes_{{\hm C}}}_{\#}P_{X^s},{\otimes_{{\hm C}}}_{\#}P_{X^t|\mathcal{Y}^s})\\=&\Lambda +d^{\ell}_{\Delta_{{\hm C},{\hm G}}}({\otimes_{{\hm C}}}_{\#}P_{X^s},{\otimes_{{\hm C}}}_{\#}P_{X^t|\mathcal{Y}^s}).
\end{split}
\end{equation}
\\

\textbf{Step 2.} we claim that   $\frac{\pi_{K+1}^t}{1-\pi_{K+1}^t}L^t_{K+1}({\hm C}\circ {\hm G})\leq  {d_{\Delta_{{\hm C},{\hm G}}}^{\ell}({\otimes_{{\hm C}}}_{\#}P_{X^s},{\otimes_{{\hm C}}}_{\#}P_{X^t|\mathcal{Y}^s})}+\max \{-\epsilon,\frac{L^t_{u,K+1}({\hm C}\circ {\hm G})}{1-\pi_{K+1}^t}-L^t_{u,K+1}({\hm C}\circ {\hm G})\}$.

\begin{equation}\label{100000}
	\begin{split}
		L^t_{u,K+1}({\hm C}\circ {\hm G})&=\int_{\mathcal{X}}\ell({\hm C}\circ {\hm G}({\mathbf{x}}),{\mathbf{y}}_{K+1}){\rm d}P_{X^t}\\&=\pi_{K+1}^t\int_{\mathcal{X}}\ell({\hm C}\circ {\hm G}({\mathbf{x}}),{\mathbf{y}}_{K+1}){\rm d}P_{X^t|{\mathbf{y}}_{K+1}}+(1-\pi_{K+1}^t)\int_{\mathcal{X}}\ell({\hm C}\circ {\hm G}({\mathbf{x}}),{\mathbf{y}}_{K+1}){\rm d}P_{X^t|\mathcal{Y}^s}.
	\end{split}
\end{equation}
We note that 
\begin{equation}\label{120}
	L_{K+1}^t({\hm C}\circ {\hm G})=\int_{\mathcal{X}}\ell({\hm C}\circ {\hm G}({\mathbf{x}}),{\mathbf{y}}_{C+1}){\rm d}P_{X^t|{\mathbf{y}}_{C+1}}.
\end{equation}
Therefore, according to (\ref{100000}) and (\ref{120}), we have
\begin{equation}\label{130}
	\begin{split}
		&L^t_{u,C+1}({\hm C}\circ {\hm G})-(1-\pi_{K+1}^t)L^s_{u,C+1}({\hm C}\circ {\hm G})\\=&\pi_{K+1}^tL^t_{C+1}({\hm C}\circ {\hm G})+(1-\pi_{K+1}^t)\int_{\mathcal{X}}\ell({\hm C}\circ {\hm G}({\mathbf{x}}),{\mathbf{y}}_{C+1}){\rm d}P_{X^t|\mathcal{Y}^s}-(1-\pi_{K+1}^t)\int_{\mathcal{X}}\ell({\hm C}\circ {\hm G}({\mathbf{x}}),{\mathbf{y}}_{C+1}){\rm d}P_{X^s}.
	\end{split}
\end{equation}
According to the definition of tensor distribution discrepancy  and the condition that the constant vector value function ${\bm g}={\mathbf{y}}_{C+1}\in \mathcal{H}_{\hm G}$, we have
\begin{equation}\label{14}
	\begin{split}
		&\left|\int_{\mathcal{X}}\ell({\hm C}\circ {\hm G}({\mathbf{x}}),{\mathbf{y}}_{C+1}){\rm d}P_{X^t|\mathcal{Y}^s}-\int_{\mathcal{X}}\ell({\hm C}\circ {\hm G}({\mathbf{x}}),{\mathbf{y}}_{C+1}){\rm d}P_{X^s}\right|\\&\leq d^{\ell}_{\Delta_{{\hm C},{\hm G}}}({\otimes_{{\hm C}}}_{\#}P_{X^s},{\otimes_{{\hm C}}}_{\#}P_{X^t|\mathcal{Y}^s}).
	\end{split}
\end{equation}

Combining (\ref{130}) and (\ref{14}), we show that
\begin{equation}\label{16}
	\begin{split}
		\pi_{K+1}^tL^t_{K+1}({\hm C}\circ {\hm G})\leq & {(1-\pi_{K+1}^t)}d^{\ell}_{\Delta_{{\hm C},{\hm G}}}({\otimes_{{\hm C}}}_{\#}P_{X^s},{\otimes_{{\hm C}}}_{\#}P_{X^t|\mathcal{Y}^s})\\+&L^t_{u,K+1}({\hm C}\circ {\hm G})-(1-\pi_{K+1}^t)L^t_{u,K+1}({\hm C}\circ {\hm G}).
	\end{split}
\end{equation}
Hence,

\begin{equation}\label{l3}
    \frac{\pi_{K+1}^t}{1-\pi_{K+1}^t}L^t_{K+1}({\hm C}\circ {\hm G})\leq  {d_{\Delta_{{\hm C},{\hm G}}}^{\ell}({\otimes_{{\hm C}}}_{\#}P_{X^s},{\otimes_{{\hm C}}}_{\#}P_{X^t|\mathcal{Y}^s})}+\max \{-\epsilon,\frac{L^t_{u,K+1}({\hm C}\circ {\hm G})}{1-\pi_{K+1}^t}-L^t_{u,K+1}({\hm C}\circ {\hm G})\}.
\end{equation}
\\
\textbf{Step 3.} It is easy to check that 
\begin{equation}\label{17}
	\begin{split}
		L^t({\hm C}\circ {\hm G})=(1-\pi_{K+1}^t)L^t_*({\hm C}\circ {\hm G})+\pi_{K+1}^t{L}^t_{K+1}({\hm C}\circ {\hm G}),
	\end{split}
\end{equation}
Finally, using (\ref{l4}), (\ref{l3}) and (\ref{17}), we have  
\begin{equation}
	~~\frac{L^t({\hm C}\circ {\hm G})}{1-\pi_{K+1}^t}\leq   {L^s({\hm C}\circ {\hm G})} +{2d^{\ell}_{\Delta_{{\hm C},{\hm G}}}({\otimes_{{\hm C}}}_{\#}P_{X^s},{\otimes_{{\hm C}}}_{\#}P_{X^t|\mathcal{Y}^s})}+\Lambda+\Delta_\epsilon.
\end{equation}
{The proof has been completed.}
\end{proof}

\begin{Proposition}
\begin{equation}
\begin{split}
\Delta&=\frac{L^t_{u,K+1}({\hm C}\circ {\hm G})}{(1-\pi_{K+1}^t)}-L^s_{u,K+1}({\hm C}\circ {\hm G}) \\& \geq \frac{\pi_{K+1}^t}{(1-\pi_{K+1}^t)}L^t_{K+1}({\hm C}\circ {\hm G})-d^{\ell}_{{\mathcal{H}_{\hm G}}}({{\hm G}}_{\#}P_{X^s},{{\hm G}}_{\#}P_{X^t|\mathcal{Y}^s}).
\end{split}
\end{equation}
\end{Proposition}
\begin{proof}
We study
\begin{equation}
	\begin{split}
		&L^t_{u,C+1}({\hm C}\circ {\hm G})-(1-\pi_{K+1}^t)L^s_{u,C+1}({\hm C}\circ {\hm G})\\=&\pi_{K+1}^tL^t_{C+1}({\hm C}\circ {\hm G})+(1-\pi_{K+1}^t)\int_{\mathcal{X}}\ell({\hm C}\circ {\hm G}({\mathbf{x}}),{\mathbf{y}}_{C+1}){\rm d}P_{X^t|\mathcal{Y}^s}-(1-\pi_{K+1}^t)\int_{\mathcal{X}}\ell({\hm C}\circ {\hm G}({\mathbf{x}}),{\mathbf{y}}_{C+1}){\rm d}P_{X^s}.
	\end{split}
\end{equation}
Note that
\begin{equation}
\left|\int_{\mathcal{X}}\ell({\hm C}\circ {\hm G}({\mathbf{x}}),{\mathbf{y}}_{C+1}){\rm d}P_{X^t|\mathcal{Y}^s}-\int_{\mathcal{X}}\ell({\hm C}\circ {\hm G}({\mathbf{x}}),{\mathbf{y}}_{C+1}){\rm d}P_{X^s}\right|\leq d^{\ell}_{{\mathcal{H}_{\hm G}}}({{\hm G}}_{\#}P_{X^s},{{\hm G}}_{\#}P_{X^t|\mathcal{Y}^s}).
\end{equation}
Hence,
\begin{equation}
\left|\Delta-\frac{\pi_{K+1}^t}{(1-\pi_{K+1}^t)}L^t_{K+1}({\hm C}\circ {\hm G})\right |\leq d^{\ell}_{{\mathcal{H}_{\hm G}}}({{\hm G}}_{\#}P_{X^s},{{\hm G}}_{\#}P_{X^t|\mathcal{Y}^s}),
\end{equation}
which implies that
\begin{equation}
\begin{split}
\Delta&=\frac{L^t_{u,K+1}({\hm C}\circ {\hm G})}{(1-\pi_{K+1}^t)}-L^s_{u,K+1}({\hm C}\circ {\hm G}) \\& \geq \frac{\pi_{K+1}^t}{(1-\pi_{K+1}^t)}L^t_{K+1}({\hm C}\circ {\hm G})-d^{\ell}_{{\mathcal{H}_{\hm G}}}({{\hm G}}_{\#}P_{X^s},{{\hm G}}_{\#}P_{X^t|\mathcal{Y}^s}).
\end{split}
\end{equation}
\end{proof}
\newpage
\section{\large{Appendix B: Network Architecture}}
\begin{table}[htbp]
% \large
\centering
\caption{The network details of generator and classifier for Office-31, Office-Home and PIE}
\label{tab:non-didit}
% p{0.75cm}<{\centering}
\begin{tabular}{p{2.5cm}|p{2.5cm}<{\centering}|p{2.5cm}<{\centering}|p{2.5cm}<{\centering}|p{2.5cm}<{\centering}}
\hline
\multirow{2}{*}{Database} & \multicolumn{3}{c|}{Generator} & {Classifier}\\
\cline{2-5}
&backbone &fully-connected 1 &fully-connected 2 &fully-connected 1\\
\hline
Office-31 &VGGNet-16 &4096 $\times$ 3000 &3000 $\times$ 2048 &2048 $\times$ 11\\
Office-Home &ResNet-50 &2048 $\times$ 3072 &3072 $\times$ 3072 &3072 $\times$ 26\\
PIE &- &1024 $\times$ 3072 &3072 $\times$ 2048 &2048 $\times$ 21\\
\hline
% fully-connected
\end{tabular}
\end{table}
It is worth noting that PIE directly provide features of $1024$ dimensions rather than images. And Leaky Relu and Batch Normalization layers are used after each layer except backbone. The parameters of backbone are not updated during the process of training. 

\begin{minipage}{\textwidth}
\makeatletter\def\@captype{table}\makeatother\caption{The network details of generator and classifier for Digit and the network details of Discriminator}
\begin{minipage}{0.5\textwidth}
\centering
\begin{tabular}{|p{6cm}<{\centering}|}
\hline
 SVHN $\rightarrow$ MNIST\\
\hline
Generator\\
\hline
Conv 3 $\times$ 3 $\times$ 64, stride 1, pad 1 \\
Conv 3 $\times$ 3 $\times$ 64, stride 1, pad 1 \\
Conv 3 $\times$ 3 $\times$ 64, stride 1, pad 1 \\
Max-pool 2 $\times$ 2, stride 2\\
Dropout 0.5\\
Gaussian noise, $\mu$=0, $\sigma$=1\\
Conv 3 $\times$ 3 $\times$ 64, stride 1, pad 1\\
Conv 3 $\times$ 3 $\times$ 64, stride 1, pad 1\\
Conv 3 $\times$ 3 $\times$ 64, stride 1, pad 1\\
Max-pool 2 $\times$ 2, stride 2\\
Dropout 0.5\\
Conv 3 $\times$ 3 $\times$ 64, stride 1, pad 1\\
Conv 3 $\times$ 3 $\times$ 64, stride 1, pad 1\\
Conv 3 $\times$ 3 $\times$ 64, stride 1, pad 1\\
Global average pool 2 $\times$ 2, stride 2\\
Fully-connected 1024 $\times$ 1024\\
\hline
Classifier \\
\hline
Fully-connected 1024 $\times$ 6\\
\hline
\end{tabular}
\end{minipage}
\begin{minipage}{0.5\textwidth}
\begin{minipage}{\textwidth}
\centering
\begin{tabular}{|p{6cm}<{\centering}|}
\hline
 MNIST $\leftrightarrow$ USPS\\
\hline
Generator\\
\hline
Conv 5 $\times$ 5 $\times$ 64, stride 1, pad 0\\
Max-pool 2 $\times$ 2, stride 2\\
Conv 3 $\times$ 3 $\times$ 128, stride 1, pad 0\\
Max-pool 2 $\times$ 2, stride 2\\
Dropout 0.5\\
\hline
Classifier \\
\hline
Fully-connected 1024 $\times$ 6\\
\hline
\end{tabular}
\vspace{0.25cm}
\end{minipage}
\begin{minipage}{\textwidth}
\centering
\begin{tabular}{|p{6cm}<{\centering}|}
\hline
 Discriminator\\
\hline
Fully-connected Input $\times$ Hiden\\
Leaky Relu \\
Dropout 0.5 \\
Fully-connected Hiden $\times$ Hiden\\
Leaky Relu \\
Dropout 0.5 \\
Fully-connected Hiden $\times$ 1\\
Sigmoid \\
\hline
\end{tabular}
\end{minipage}
\end{minipage}
\vspace{0.5cm}
\end{minipage}
The design of the network architectures of SVHN $\rightarrow$ MNIST and MNIST $\leftrightarrow$ USPS are based on \cite{shu2018a} and \cite{saito2018open} respectively. For SVHN $\rightarrow$ MNIST, Leaky Relu and Batch Normalization layers are used after convolution and fully-connected layers. For MNIST $\leftrightarrow$ USPS, Leaky Relu and Batch Normalization layers are used after the convolution and fully-connected layers of the generator. For the network of discriminator, \textit{Input} is the output of the generator. The number of \textit{Hiden} is set as 500 except Office-31, and \textit{Hiden} is set as 1000 for Office-31.

\bibliographystyle{IEEEtran}
\bibliography{reference2}